\documentclass[letterpaper, 10 pt, conference]{template/ieeeconf}  

\IEEEoverridecommandlockouts                              

\usepackage{amsmath} 
\usepackage{amssymb}  
\usepackage{graphicx}
\usepackage{hyphenat}
\usepackage[table]{xcolor}
\usepackage[font=small,belowskip=0pt, aboveskip=6pt]{caption}
\usepackage{subcaption}

\usepackage{multirow}
\usepackage{booktabs,amsfonts,dcolumn}
\usepackage{tabularx}
\usepackage{makecell}
\usepackage{hyperref}
\usepackage{comment}
\usepackage{diagbox}
\usepackage{multirow}
\usepackage{calrsfs}

\usepackage{xspace}
\usepackage[group-separator={,}]{siunitx}

\DeclareMathAlphabet{\pazocal}{OMS}{zplm}{m}{n}

\long\def\invis#1{}

\newcommand\fig[1]{Fig. \ref{#1}}

\usepackage{fancyhdr}
\fancypagestyle{withfooter}{
  
  \fancyhead[L]{}
  \fancyhead[R]{}
  \fancyfoot[C]{\footnotesize Presented at the 2025 IEEE ICRA Workshop on Field Robotics}
}

\makeatletter
\DeclareRobustCommand\onedot{\futurelet\@let@token\@onedot}
\def\@onedot{\ifx\@let@token.\else.\null\fi\xspace}

\def\etal{\emph{et al}\onedot}
\makeatother

\title{\LARGE \bf
Mapping the Catacombs:\\
An Underwater Cave Segment of the Devil's Eye System
}

\author{ Michalis Chatzispyrou$^{a*}$, Luke Horgan$^{b*}$, Hyunkil Hwang$^{a*}$, Harish Sathishchandra$^{b*}$, 	 \\
Monika Roznere$^c$, Alberto Quattrini Li$^d$, Philippos Mordohai$^b$, Ioannis Rekleitis$^a$%
\thanks{$^*$ The first four authors have contributed equally to this work and are listed in alphabetical order.}
\thanks{$^a$University of Delaware, Newark, DE, USA, {\tt\small \{michalis,hkhwang,yiannisr\}@udel.edu.}}%
\thanks{$^b$Stevens Institute of Technology, Hoboken, NJ, USA, {\tt\small \{lhorgan,hsathish,pmordoha\}@stevens.edu}}%
\thanks{$^c$Binghamton University, Binghamton, NY, USA {\tt\small mrozner1@binghamton.edu}}
\thanks{$^d$Dartmouth College, Hanover, NH, USA {\tt\small alberto.quattrini.li@dartmouth.edu}}
\thanks{This research has been supported in part by the National Science Foundation under grants 1943205, 2024541, 2024653, and 2024741. The authors are grateful for the assistance of Greg St Amand, Per Normark, and Evelyn Unti during the data collection. Furthermore, the authors would like to acknowledge the help from Woodville Karst Plain Project (WKPP), El Centro Investigador del Sistema Acuífero de Quintana Roo A.C. (CINDAQ),  Global Underwater Explorers (GUE), Ricardo Constantino, and Project Baseline over the years in providing access to challenging underwater caves and mentoring us in underwater cave mapping. The authors are also grateful for equipment support by Halcyon Dive Systems, Teledyne FLIR LLC, and KELDAN GmbH lights.}
}

\begin{document}

\maketitle
\thispagestyle{withfooter}
\pagestyle{withfooter}
\begin{abstract} 
This paper presents a framework for mapping underwater caves. Underwater caves are crucial for fresh water resource management, underwater archaeology, and hydrogeology. 
Mapping the cave's outline and dimensions, as well as creating photorealistic 3D maps, is critical for enabling a better understanding of this underwater domain. 
In this paper, we present the mapping of an underwater cave segment (the catacombs) of the Devil's Eye cave system at Ginnie Springs, FL. We utilized a set of inexpensive action cameras in conjunction with a dive computer to estimate the trajectories of the cameras together with a sparse point cloud. The resulting reconstructions are utilized to produce a one-dimensional retract of the cave passages in the form of the average trajectory together with the boundaries (top, bottom, left, and right). The use of the dive computer enables the observability of the z-dimension in addition to the roll and pitch in a visual/inertial framework (SVIn2).  In addition, the keyframes generated by SVIn2 together with the estimated camera poses for select areas are used as input to a global optimization (bundle adjustment) framework -- COLMAP --  in order to produce a dense reconstruction of those areas. The same cave segment is manually surveyed using the MNemo V2 instrument, providing an additional set of measurements validating the proposed approach. It is worth noting that with the use of action cameras, the primary components of a cave map can be constructed. Furthermore, with the utilization of a global optimization framework guided by the results of VI-SLAM package SVIn2, photorealistic dense 3D representations of selected areas can be reconstructed. 
\end{abstract}

\section{Introduction}
Underwater caves are crucial for groundwater flow in karst topographies, supplying freshwater to almost 25\% of the world's population \cite{karstbook}. 
In the US, 23\% of freshwater withdrawals in 2000 came from bedrock aquifers, with Florida's aquifer having the highest withdrawal rate \cite{Covington2011}. Given its importance, a project started in $1987$ in the Woodville Karst Plain (WKPP), which resulted in the exploration of over $34$ miles of cave systems in Florida~\cite{WKPP,WKPP2}, proving the cave system to be the longest in USA~\cite{WKPP1}.
Furthermore, caves provide a unique preservation environment for archaeological secrets~\cite{CampbellBook,campbell2018introduction,gonzalez2008arrival}, yielding to major historical finds across the globe, such as Mexico~\cite{arroyo2015underwater,rissolo2015novel,de2015ancient,macdonald2020paleoindian}, Florida~\cite{faught2004underwater}, Greece~\cite{galanidou2020greece},  and Philippines~\cite{peterson2017archaeological}. 

Thus, mapping underwater caves is important~\cite{kambesis2007importance}. Currently, cave divers dive in these extreme environments and map caves with basic tools, such as knotted lines and compass, to obtain the topology of a cave~\cite{heinerth2019exploration}.
It is very dangerous even for highly skilled people~\cite{Exley1977}, with more than $350$ deaths since $1969$ in the USA alone.  These environments are still quite challenging for an underwater robot to explore given the absence of any localization or communication infrastructure. Hence, the community would benefit from progress in automated technology tools that can accurately map such environments, but are also easy to use to reduce the cognitive load and preserve the safety of divers. 
\begin{figure}[t]
     \centering
     {\includegraphics[width=0.9\columnwidth]{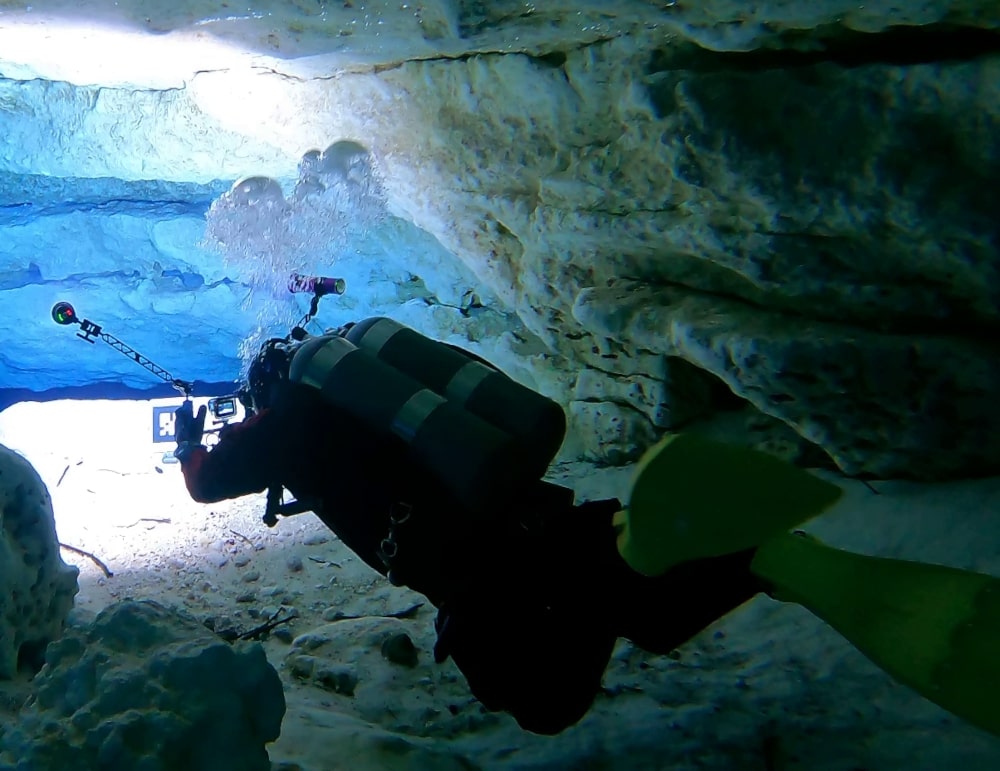}}
     \caption{GoPro setup deployed inside a cave, Ginnie Springs, FL. A fiducial tag can be seen, used for estimating ground truth. }
     \label{fig:gopro}
 \end{figure}

Mapping underwater caves presents several challenges. LiDAR and RGB-D sensors that provide dense point cloud cannot be used underwater because of the light absorption. The most common mapping sensor used underwater is sonar~\cite{mcconnell2022perception}, however, measurements can be inaccurate due to comparatively large beam width and multipath effects. Cameras are promising due to their low cost and amount of scene information it can capture, but to use them effectively, it is fundamental to overcome limited visibility, color absorption, hazing, absence of natural light, etc.\ \cite{islam2024computer}.

Some work produced good mapping results with cameras, e.g., using light to infer the surrounding  structures~\cite{WeidnerICRA2017}; integrating camera, sonar, and IMU to leverage their complementarity~\cite{RahmanIJRR2022}. 
Still, a large dense map of an underwater cave has not been fully demonstrated yet, especially using only tools available to cave divers.

This paper presents a framework for mapping underwater caves using inexpensive action cameras and a dive computer, carried by the divers. We integrate a visual-inertial SLAM framework, which provides a reduced set of images that are informative from the state estimation perspective, together with a structure-from-motion global bundle adjustment to improve geometric consistency and achieve a denser reconstruction. We use datasets collected by cave divers -- see, e.g., Fig.~\ref{fig:gopro} -- over multiple passes to verify the quality of the reconstruction of the proposed framework and compare it with the manual survey that cave divers perform. The dataset has a long duration of 81 minutes and a covered distance of more than 1km.

Besides presenting this framework, this paper contribution is twofold. First, we provide a unique dataset on a cave system that contains several branches and loops and that we will make public. Second, we discuss open problems from the lessons learned in reconstructing a segment of that cave system, towards advancing technology for divers and developing robust underwater robots.

\section{Related Work}\label{sec:related}
While the literature is vast on 3D mapping, ranging from photogrammetry \cite{aicardi2018recent} to simultaneous localization and mapping \cite{placed2023survey} and more recently Neural Radiance Field/Gaussian Splatting~\cite{mualem2024gaussiansplashing}, we discuss here work specifically related to underwater cave mapping, highlighting the current state of the art and main challenges and gaps.

Traditionally, scuba divers performed the topological mapping of underwater caves using knotted lines or measuring tape for distance, a compass for orientation, and a (water) depth sensor/dive computer for recording the depth. Cartographers would then use the survey information to create depictions of the caves. This process is valuable, as it provides an understanding of how caves expand, but is prone to error. The cave diving community looks forward to technological progress making these surveys easier \cite{heinerth2019exploration}. In this paper, we integrate surveying methods used by the cave diving community with 3D reconstruction pipelines.

There has long been robot technology that can support cave surveying, even if they are expensive and require expertise in processing the collected data. 
In 2007, Stone Aerospace presented DEPTHX (DEep Phreatic THermal eXplorer)~\cite{Stone2007}, a teleoperated robot, equipped with an IMU, two
depth sensors, a DVL, and 54 narrow single-beam
echosounders. DEPTHX was used to map a vertical shaft filled with water~\cite{gary20083d}.
Later, the same company developed Sunfish \cite{richmond2018sunfish}, an autonomous underwater vehicle with a multibeam sonar, an underwater dead-reckoning system based on a fiber-optic gyroscope (FOG) IMU, DVL, and pressure-depth sensors. Sunfish was deployed inside some caves in Florida. However, sonar-based reconstruction can be inaccurate in caves due to wide beam width and multipath effects \cite{guerneve2018three}.

The advancements in camera technology and their miniaturization resulted in a broadening of their use during cave exploration by divers. However, full 3D reconstruction of underwater caves remains an open problem.
Some attempts that specifically considered ease of use for divers included a solution that used the boundary between the scuba dive light and shadow to enhance feature matching, tracking, and 3D reconstruction~\cite{WeidnerICRA2017}.
A visual-inertial SLAM system, called SVIn2 \cite{RahmanICRA2018,RahmanIROS2019a,RahmanIJRR2022}, integrated stereo cameras, an IMU, a depth sensor, and a profiling sonar in a tightly-coupled optimization framework to improve the state estimation reliability in underwater scenarios. Experiments were carried out using a custom stereo rig with such sensors~\cite{RahmanOceans2018}, as well as GoPro cameras that have an IMU \cite{JoshiICRA2022}. The integration of the two systems, light and visual-inertial SLAM, system resulted in a denser and accurate contour-based reconstruction \cite{RahmanIROS2019b}. SVIn2 is the base of our reconstruction pipeline, given its robustness.

Photogrammetry can help in recording marine cave assemblages \cite{pulido2023photogrammetry} and achieving high-resolution 3D models \cite{calantropio2024underwater}, but it presents several challenges in achieving highly accurate results due to low visibility, light degradation, and absence of well-defined landmarks. Recently, an approach that combines VIO and a structure-from-motion pipeline demonstrated abilities to reconstruct sections of a cave \cite{lago2024visual}. Similarly, we integrate the visual SLAM pipeline with a global bundle adjustment system, though we also feed reliable keyframes to achieve more accurate reconstructions.

Generally, compared to other domains, there is a lack of underwater cave datasets. Only one is currently available for general use \cite{mallios2017underwater}, though there is another for benchmarking VIO/SLAM systems~\cite{JoshiIROS2019}. 
A dataset for cave visual segmentation has also been recently released \cite{AbdullahICRA2024}. We will contribute a new cave dataset that comprises a large system with several branches and loops.

\section{Proposed Framework}
\subsection{Experimental setup}
Three GoPro\textsuperscript{TM} Hero9 black action cameras were used. The cameras were rigidly mounted on an aluminum frame with handles; see \fig{fig:gopro}, and they were carefully positioned to produce a wide, partially overlapping field of view. In particular, the left and right cameras were aligned horizontally but facing approximately thirty degrees off center in opposing directions, while the center camera was pointed approximately 30 degrees upward. Two Keldan\textsuperscript{TM} video lights were connected to the handles to generate the required illumination, as there is no natural light inside the cave. One Shearwater\textsuperscript{TM} Perdix2 AI dive computer (carried often by divers) was utilized to record the depth of the water at the position of the diver, who operated the camera rig.

The video clips from the camera are converted into a ROS1 bagfile using the approach proposed by Joshi \etal~\cite{JoshiICRA2022,goproROS}. This framework encodes the video (at 30 fps) and the IMU (100 Hz) datastream as ROS1 messages with synchronized timestamps. The resulting bagfile can be played back and used as input to a number of VIO/VI-SLAM packages; in our case SVIn2~\cite{RahmanIJRR2022} is used.  

\begin{figure}[h]
    \centering
    \begin{tabular}{cc}
         \fbox{\includegraphics[height=0.15\textheight, trim={0in, 0in, 0in, 0in},clip]{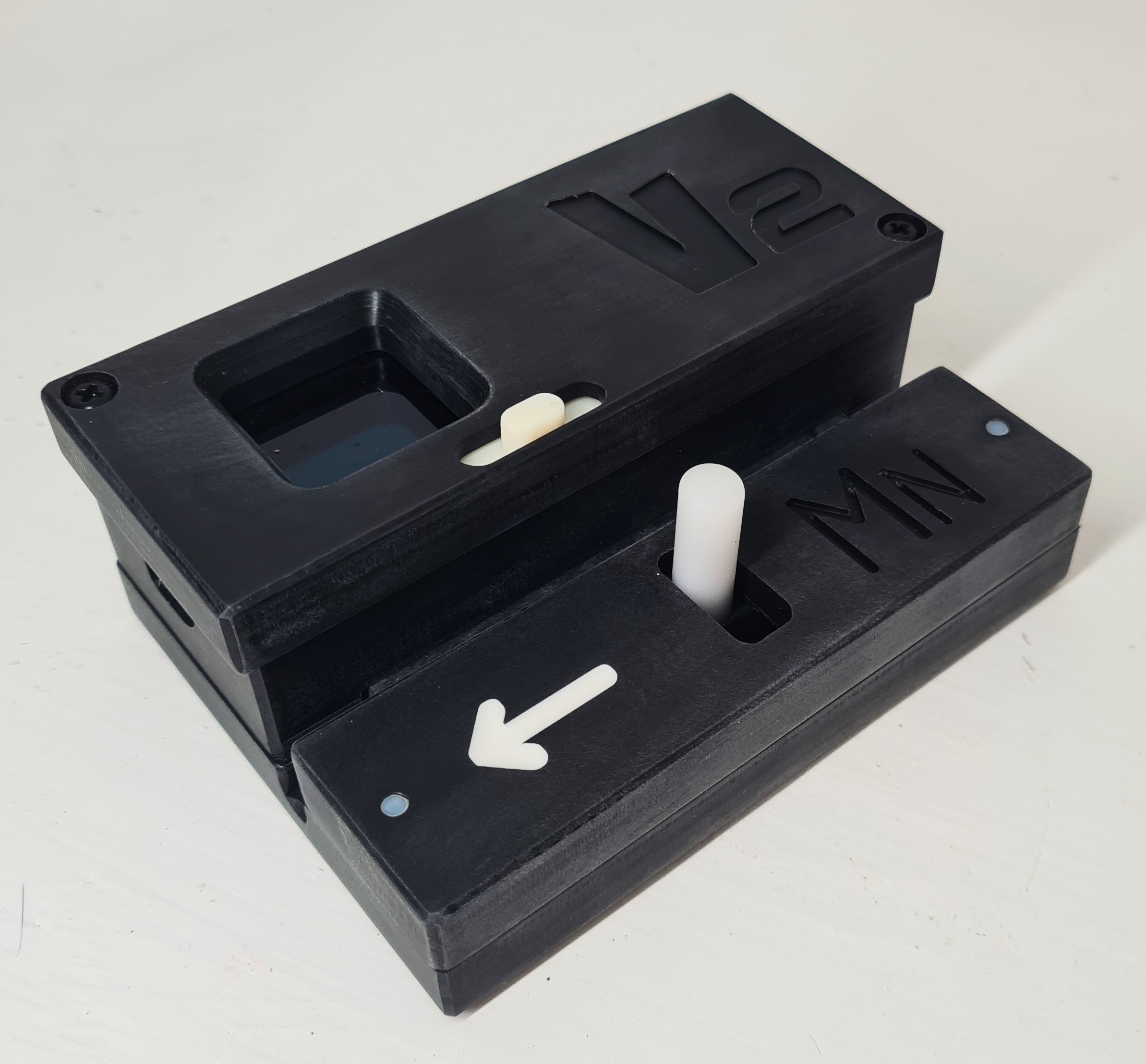}}&
         \fbox{\includegraphics[height=0.15\textheight, trim={0in, 0in, 0in, 0in},clip]{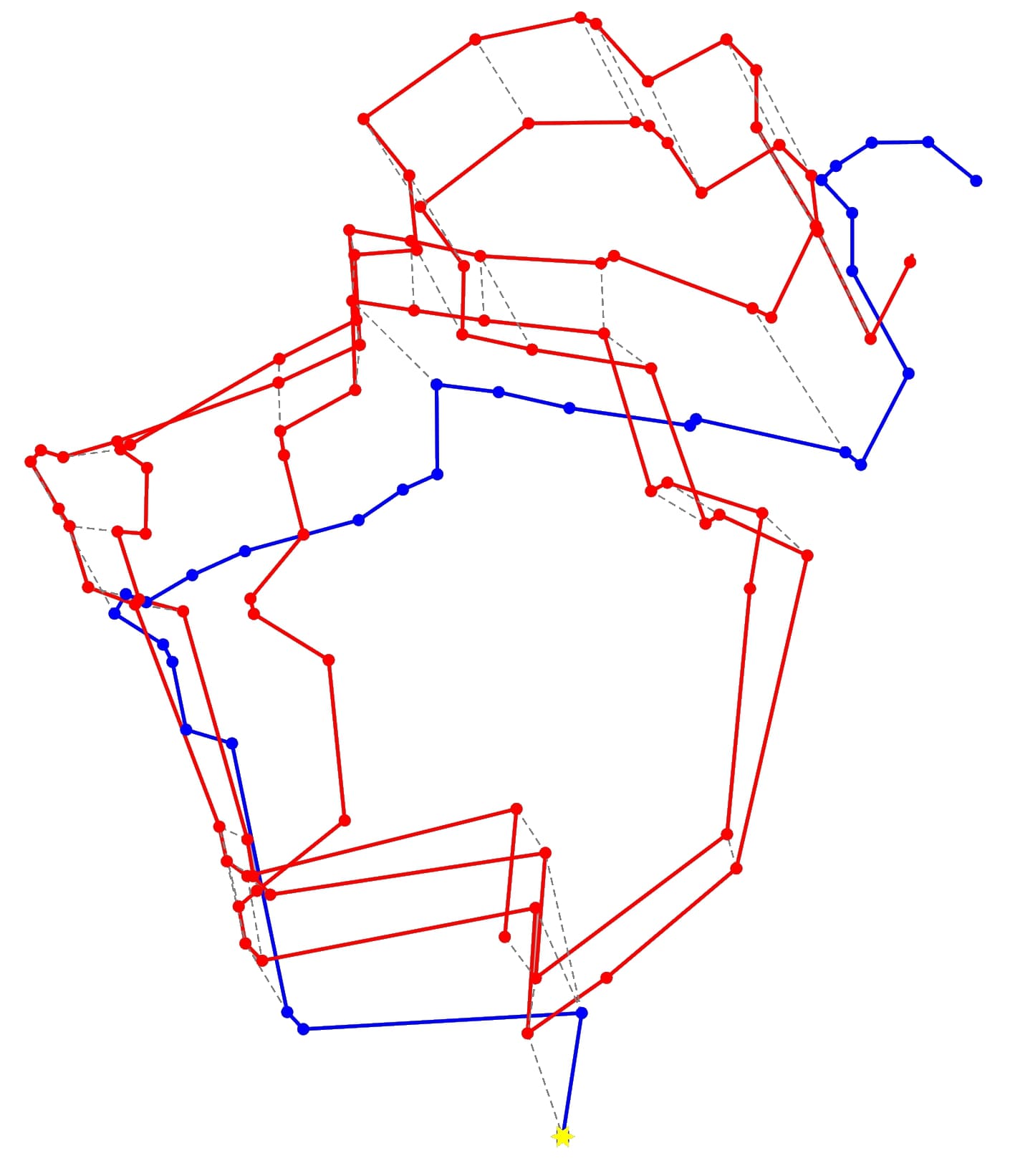}}
   \end{tabular}
    \caption{MNemo~\cite{mnemo}: underwater cave survey data acquisition device. Raw data collected by MNemo with loops manually selected for least square loop closure in the Ariane software~\cite{ariane}.}
    \label{fig:mnemo}
\end{figure}

In a separate deployment, the MNemo survey data acquisition system~\cite{mnemo} (see \fig{fig:mnemo} left) is used to record data pertaining to the deployed caveline. MNemo records depth and azimuth at each endpoint of a segment and also the length of the segment. The right image in \fig{fig:mnemo} presents the raw data as recorded from MNemo. Please note that, during deployment, the data acquisition was interrupted due to operator error (blue line) and re-started (red line). The software Ariane~\cite{ariane} was used to calculate loops (light grey dashed lines) and then produce a stick map of the caveline; see \fig{fig:TrajMaps} for the processed map. 

\subsection{Visual-Inertial SLAM -- SVIn2}
We utilize the Visual Inertial SLAM framework SVIn2~\cite{RahmanIJRR2022} for extracting the camera trajectory and a sparse point cloud. Although the detailed description is beyond the scope of this paper, the main points of this framework are described next.

SVIn2 is a tightly-coupled optimization framework that integrates IMU, camera, water pressure sensor, and sonar, by jointly minimizing error terms for each sensor. SVIn2 operates with a frontend, which provides a trajectory based on the optimization in a local map, and with a backend that supports loop closure for global trajectory optimization of keyframes. A new keyframe is one that maintains a certain overlap with a previous keyframe. A subsequent extension \cite{JoshiICRA2022} uses the optimized camera poses from the backend to update, in the frontend, the point clouds visible from each camera pose. 

At the end, our proposed pipeline extracts the camera trajectory composed of keyframes and corresponding point clouds used in the next steps of the pipeline.

\subsection{Absolute Water Depth Correction}
\begin{figure}[h]
    \centering
    \begin{tabular}{lc}
    \begin{subfigure}{0.22\textwidth}
         {\includegraphics[height=0.125\textheight]{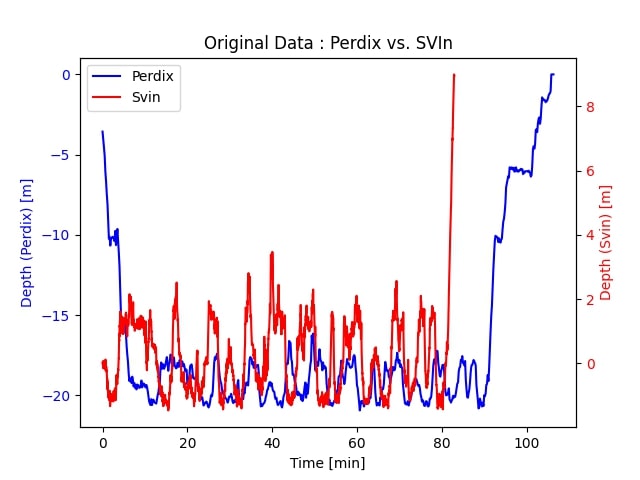}}
         \caption{} \label{depthProcessA}
   \end{subfigure}&
   \begin{subfigure}{0.22\textwidth}
         {\includegraphics[height=0.125\textheight, trim={0.2in, 0in, 0in, 0.3in},clip]{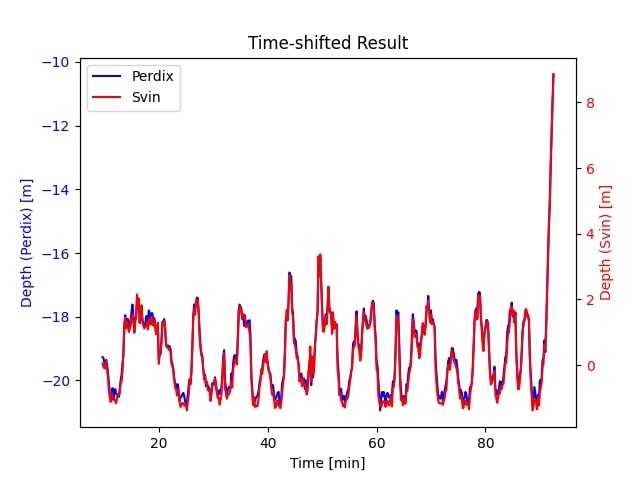}}
         \caption{} \label{depthProcessB}
   \end{subfigure}\\
    \begin{subfigure}{0.22\textwidth}
         {\includegraphics[height=0.125\textheight, trim={0.2in, 0in, 0.5in, 0.3in},clip]{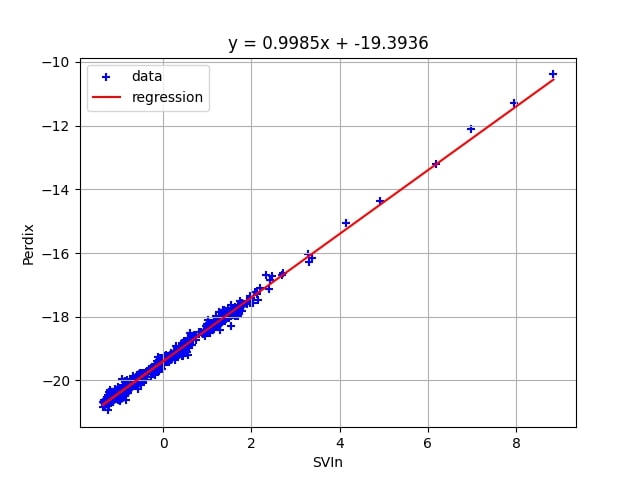}}
         \caption{} \label{depthProcessC}
   \end{subfigure}&
   \begin{subfigure}{0.22\textwidth}
         {\includegraphics[height=0.125\textheight, trim={0.2in, 0in, 0.5in, 0.3in},clip]{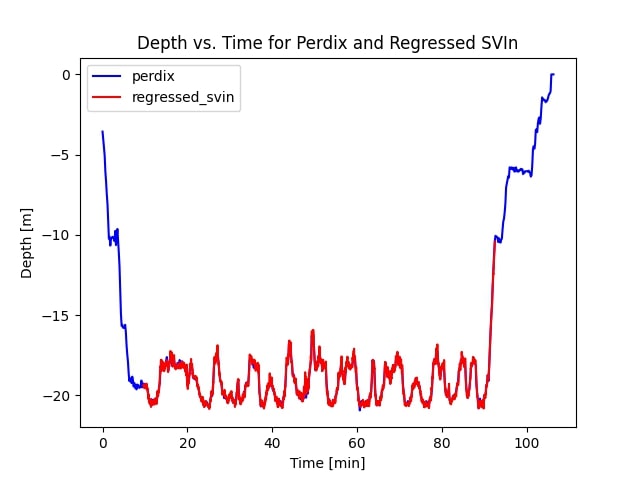}}
         \caption{} \label{depthProcessD}
    \end{subfigure}\\
    \begin{subfigure}{0.22\textwidth}
         {\includegraphics[height=0.125\textheight, trim={0.2in, 0in, 0.5in, 0.3in},clip]{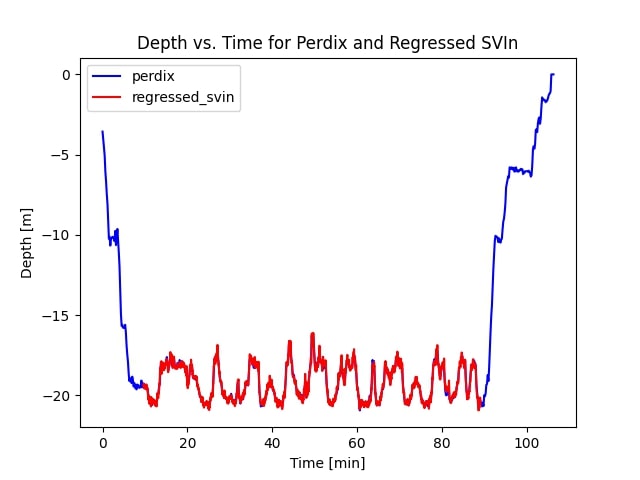}}
         \caption{} \label{depthProcessE}
   \end{subfigure}&
   \begin{subfigure}{0.22\textwidth}
         {\includegraphics[height=0.125\textheight, trim={0.2in, 0in, 0.5in, 0.3in},clip]{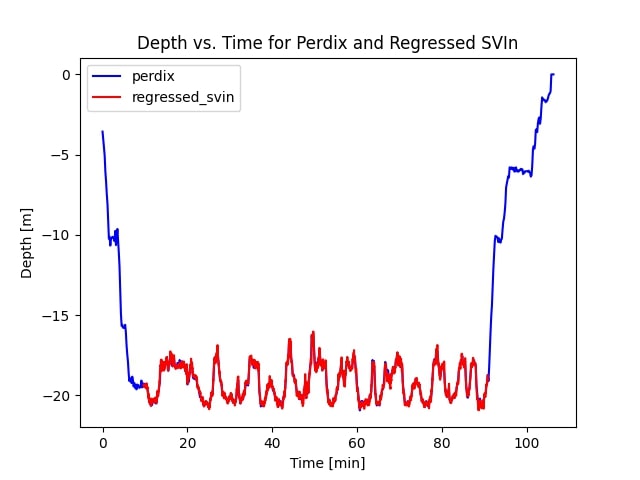}}
         \caption{} \label{depthProcessF}
    \end{subfigure}
    \end{tabular}
    \caption{(a) The SVIn2 z-coordinates (red) and the dive computer depth (blue) before synchronization for the left GoPro dataset. (b) The two time series shifted to a common time but at different scales. (c) Linear regression between the SVIn2 and the Perdix data. (d) The depth estimates time-shifted and with adjusted depth; in red, the SVIn2 produced trajectory; in blue, the data from the dive computer, for the left GoPro dataset. (e),(f) The depth estimates time-shifted and with adjusted depth; in red, the SVIn2 produced trajectory; in blue, the data from the dive computer, for the right and center GoPro datasets respectively.}
    \label{fig:depthProcess}
\end{figure}
The dive computer (Perdix) provides precise depth measurements throughout the entire dive. In contrast, the SVIn2 trajectory starts from a random location and provides relative positioning based on the starting point, with absolute positions remaining unknown~\cite{hesch2014camera,yang2019observability}. In addition, the two signals are sometimes acquired in different time zones, resulting in significant time drift. Therefore, we adjusted the SVIn2 start time to match the Perdix start time to reduce the time drift to some extent; see \fig{fig:depthProcess}(a) for the asynchronous data in the Catacombs-Left dataset. Moreover, the dive computer saves depth data every 10 seconds at a frequency of 0.1 Hz, which is much slower than the frequency of SVIn2 based on the selected keyframes; in our case 2.8 Hz (Catacombs-Left), 3.8 Hz (Catacombs-Right), and 4 Hz (Catacombs-Center).

\begin{figure*}[ht]
    \centering
    \begin{tabular}{lccc}
         \hspace{-0.05in}\fbox{\includegraphics[height=0.15\textheight, trim={1.1in, 1.1in, 0.9in, 0.9in},clip]{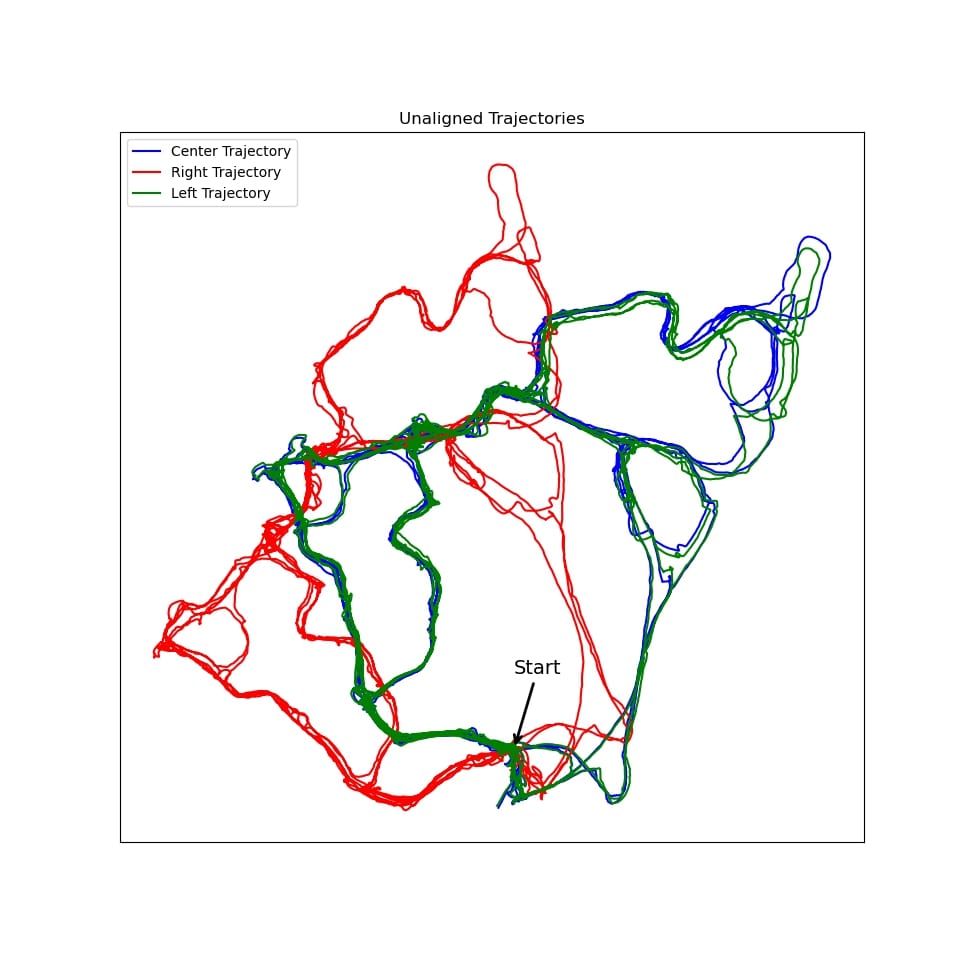}}&
         \hspace{-0.05in}\fbox{\includegraphics[height=0.15\textheight, trim={1.0in, 1.0in, 0.9in, 0.9in},clip]{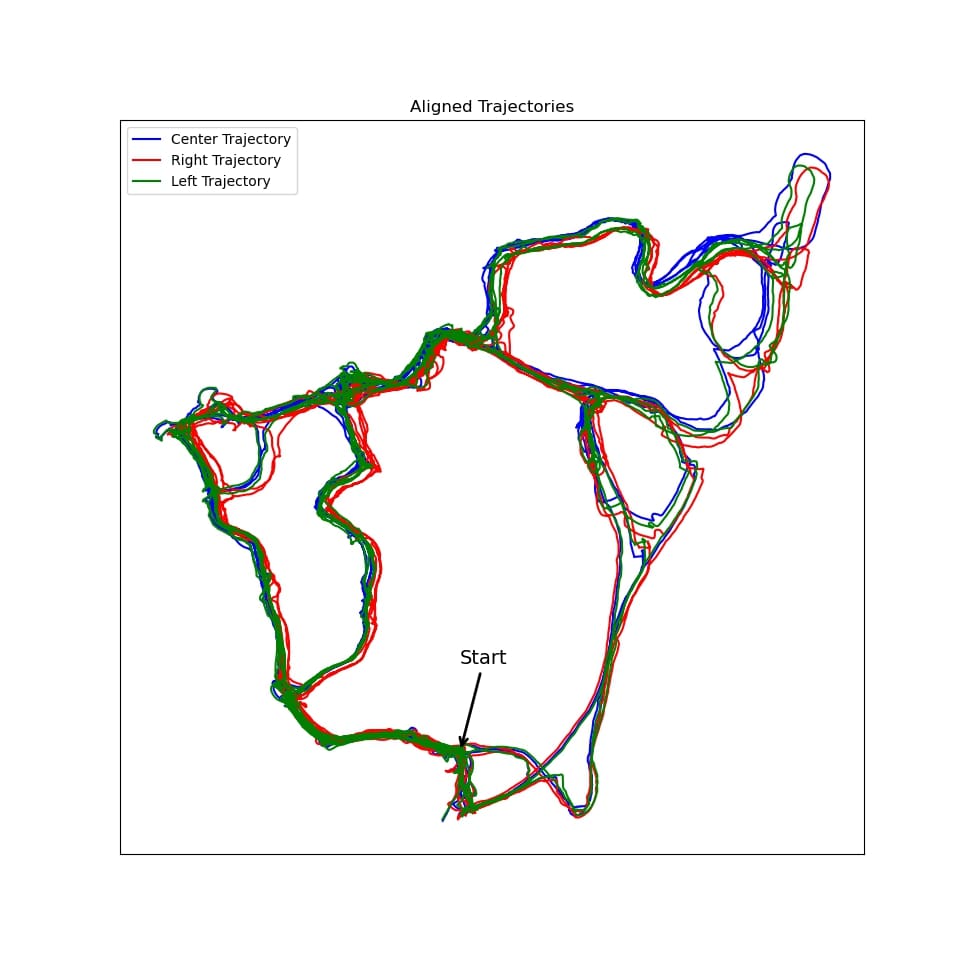}}&
         \hspace{-0.05in}\fbox{\includegraphics[height=0.15\textheight, trim={0in, 0in, 0in, 0in},clip]{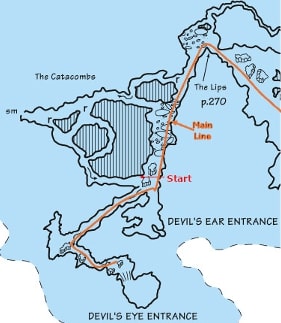}}&
         \hspace{-0.05in}{\includegraphics[height=0.15\textheight, trim={0in, 0in, 0in, 0in},clip]{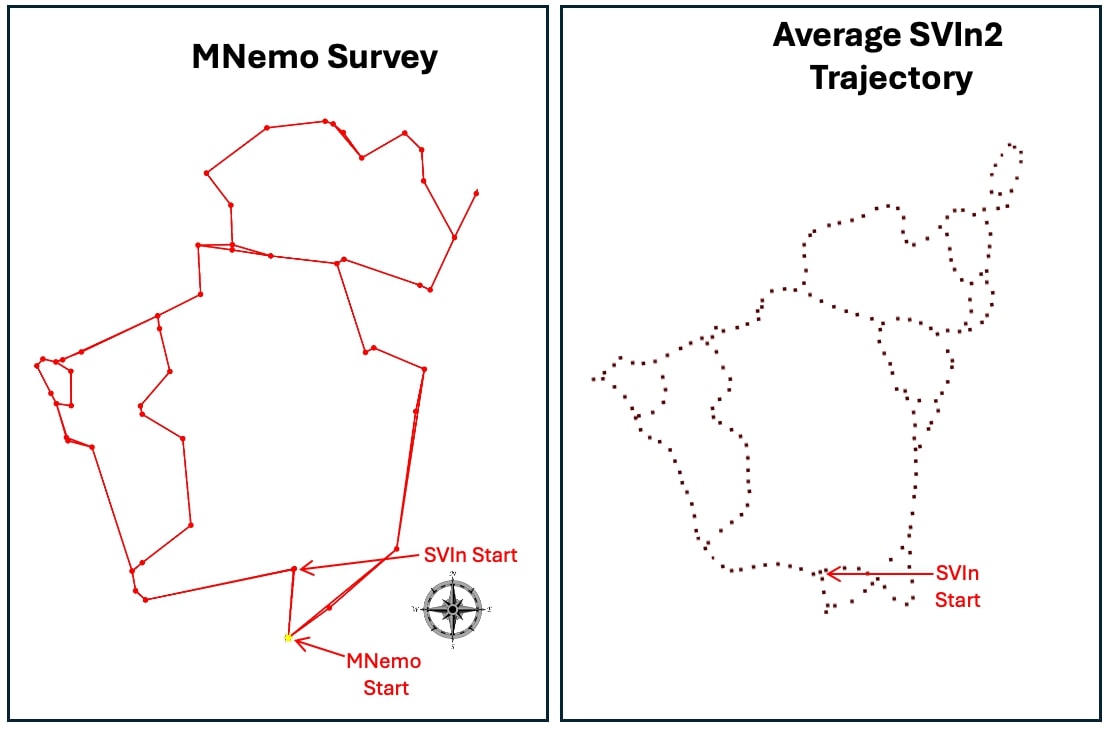}}
    \end{tabular}
    \caption{From the left: The un-aligned trajectories produced from SVIn2 for the left (green), center (blue), and right (red) GoPro. The aligned trajectories. Segment of the cave map with the main line overlayed and the start of the deployment marked. The results from the manual survey using MNemo2~\cite{mnemo} survey tool and Ariane~\cite{ariane} software. Last on the right, the average camera trajectory from SVIn2.}
    \label{fig:TrajMaps}
\end{figure*}

For every dataset, we already aligned the SVIn2 and Perdix data, thus they have the same start time. In the first step, the depth data is interpolated to 100 Hz to match the same frequency. These interpolated signals were used to determine the time shift values between them by calculating the cross-correlation of the two datasets; see \fig{fig:depthProcess}(b) where the Catacombs-Left time sequence is shifted to align with the Perdix time sequence, whereas they still have different scales (Perdix in blue and SVIn2 in red). The following step is to calculate the depth shift values between the two data using linear regression; see \fig{fig:depthProcess}(c), which indicates the input data (blue) and the regression result (red). Finally, the SVIn2 depth data are shifted to the real depth based on the values calculated by regression; see \fig{fig:depthProcess}(d) where the SVIn2 and Perdix data are aligned to the same depth scale and time in the Catacombs-Left dataset. The same approach is implemented for the Catacombs-Right (see \fig{fig:depthProcess}(e)) and Catacombs-Center (see \fig{fig:depthProcess}(f)) datasets. For all three datasets, the GoPro (SVIn2) time was shifted by 9.64 min relative to the initial Perdix time. The depth offsets were 19.39 m (Catacombs-Left), 19.4 m (Catacombs-Right), and 19.36 m (Catacombs-Center), respectively. These results reflect the experimental setup, as the three GoPro cameras were aligned in the rig, had similar depths and were set to the same time zone.

\subsection{Trajectory Alignment}\label{sec:traj_corr}
For transforming the three trajectories to the same coordinate frame, we utilize a calibration target that was placed at the begining of the deployment; see the first image of \fig{fig:denseReco} for a reconstruction of the target inside the cave passage. The  aprilgrid package~\cite{aprilgrid} is employed to detect the target in each of the datasets and the relative pose $T^{C_i}_{m}$ between the camera at time i ($C_i$) and the target ($m$). 
Consequently, by employing the pose of the camera ($P^W_{C_i}$) in world coordinates, we can estimate the pose of the target in world coordinates as $~^{C_i}P^W_{m}=P^W_{C_i}\cdot T^{C_i}_{m}$. 

First we estimate the average pose for the target as $\overline{P}=\sum_{i=1\ldots N} \frac{~^{C_i}P^W_{m}}{N}$. Then, the distance ($D_i$) is calculated between the target $~^{C_i}P^W_{m}$ for each camera $i$ that viewed the target and the average pose of the target ($\overline{P}$) as $D_i=\lVert~^{C_i}P^W_{m}-\overline{P}^j\rVert$. We calculate the standard deviation $\sigma_D$ of all the distances and then remove all targets with distances more than one $\sigma_D$ from the mean. Then, we filter outliers based on the orientation estimates, which are more noisy. From the rotation matrix of the pose of the target in the global coordinate system we extract the angles around the principal axes (roll, pitch, yaw) then we remove all the targets with at least one angle (roll, pitch, or yaw) that falls outside one standard deviation. From the remaining (inlier) targets we estimate an average pose. If there are two trajectories (A and B) the result is two poses $P^{W_A}_{m}$ and  $P^{W_B}_{m}$ for the same target in two different coordinate systems. The coordinate transformation from World frame B to world frame A is $T^{W_B}_{W_A}=P^{W_A}_{m}\cdot (P^{W_B}_{m})^{-1}$ and brings trajectory B in the coordinate frame of trajectory A. 

\subsection{Dense Reconstruction -- COLMAP}
Our pipeline includes a stucture-from-motion system, COLMAP~\cite{colmap_sfm}, run on the keyframes using as prior corresponding poses, identified by SVIn2. This step refines the geometric consistency of the camera poses given that all collections of images are used. The output will be a sparse point cloud.
We then densify that point cloud with a depth map estimation using patch-match stereo with per-pixel view selection and fusion.

To obtain the mesh reconstruction, we use a screened Poisson reconstruction, where an implicit function indicator is computed to determine whether a point is inside or outside the model \cite{kazhdan2013screened}.

\section{Experimental Results}
\subsection{Dataset}
The catacombs section of the Devil's Eye cave system is a narrow passage to the left of the main line as we are entering the cave. The area is characterized by several interconnected narrow tunnels, some of which are accessible only in side-mount configuration\footnote{In cave diving the passages are characterized as back-mount, where the diver carries double scuba tanks on their back; side-mount, where the tanks are slunk on the diver's side, thus creating a narrower vertical profile; and no-mount, where the diver pushes their tanks in front of them, while breathing from them.}; see \fig{fig:TrajMaps} center for a depiction of the area. Please note that the cave map presented in the center of \fig{fig:TrajMaps} is not metrically accurate but provides a generic outline of the area. The catacombs are not marked with a permanent line, however, for the experiments presented here, one was temporarily installed. During a single deployment of one hour, forty-six minutes at an average depth of seventeen meters, the area was mapped using the three cameras described in the experimental setup section. Multiple loops were performed as can be seen in \fig{fig:TrajMaps}, covering all accessible parts of the catacombs. At the entrance of the passage a calibration target was placed and at the start of the deployment a short calibration motion sequence was performed. In total, data collection lasted for 81 minutes and the total estimated trajectory was more than 1km. 

A second dive was performed for manually surveying the caveline deployed inside the catacombs. We started at the mainline (see \fig{fig:TrajMaps} center) and followed the line into the catacombs; see the second from right of \fig{fig:TrajMaps} where the MNemo start is marked and also, where the VI-SLAM process started. The data collection process was hampered by the current (flow) that is ever-present in the Devil's Eye cave system. Small orientation errors resulted in deviation of the line segments; see \fig{fig:mnemo}, however, the Ariane~\cite{ariane} software optimized the marked loop closures resulting in the map presented in \fig{fig:TrajMaps}.  

\begin{figure*}[h]
    \centering
    \begin{tabular}{lccc}
         \hspace{-0.05in}\fbox{\includegraphics[height=0.14\textheight, trim={0in, 0in, 0in, 0in},clip]{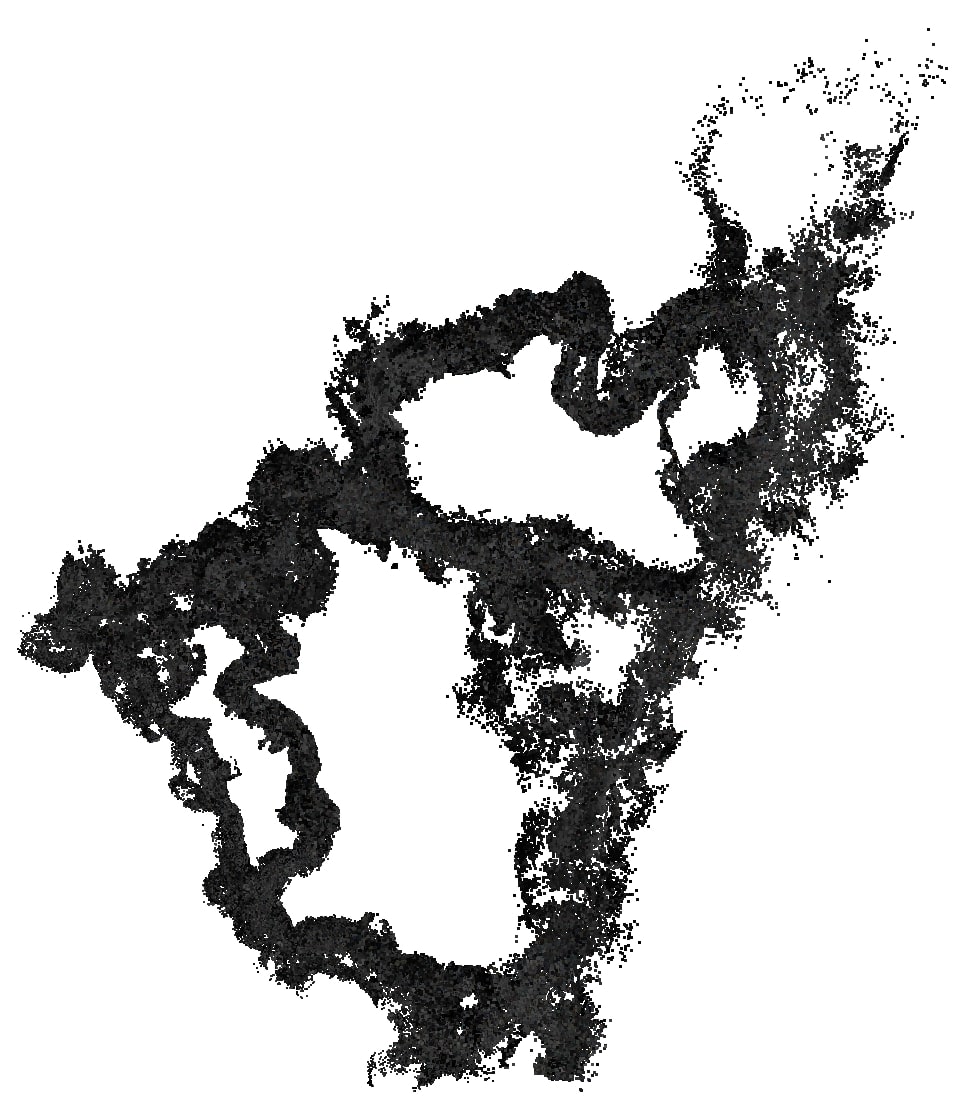}}&
         \hspace{-0.05in}\fbox{\includegraphics[height=0.14\textheight, trim={0in, 0in, 0in, 0in},clip]{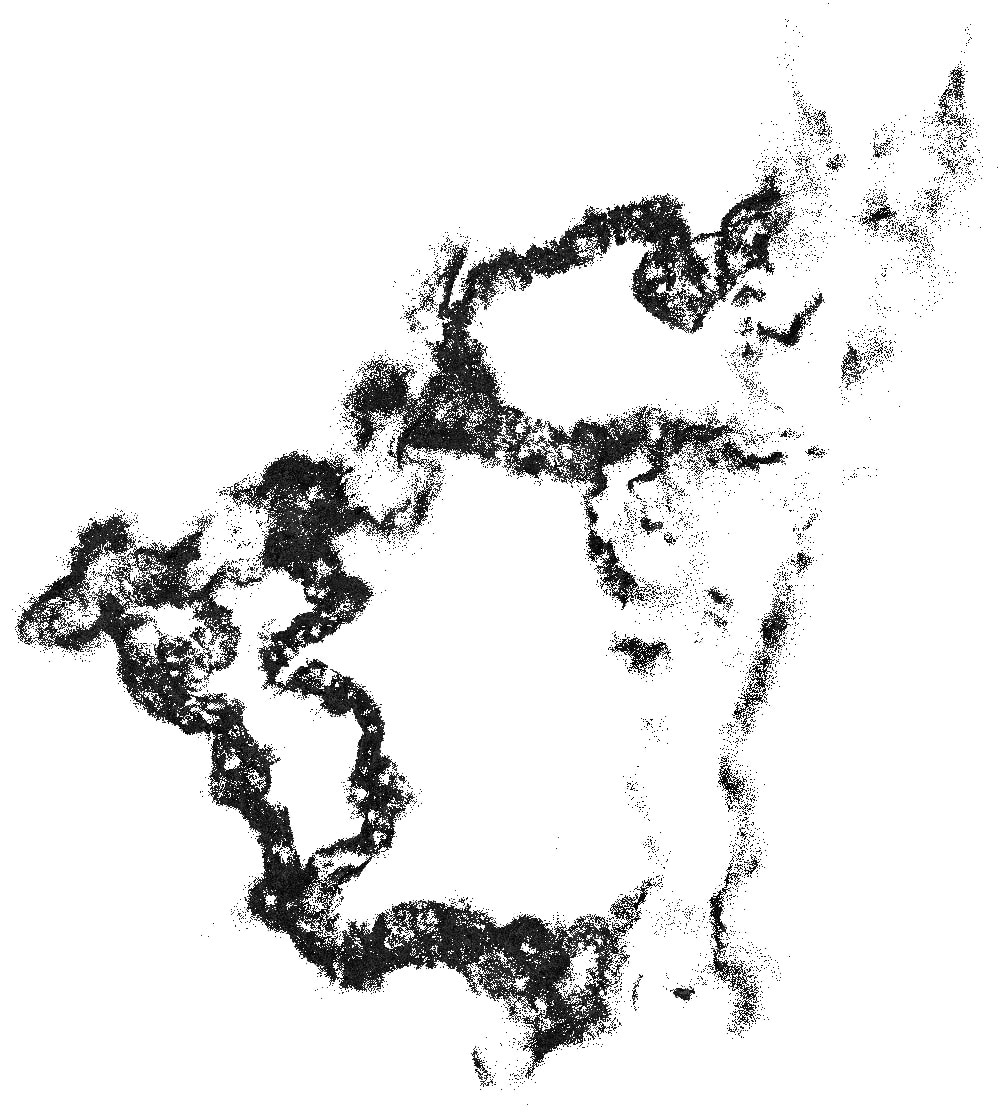}}&
         \hspace{-0.05in}\fbox{\includegraphics[height=0.14\textheight, trim={0in, 0in, 0in, 0in},clip]{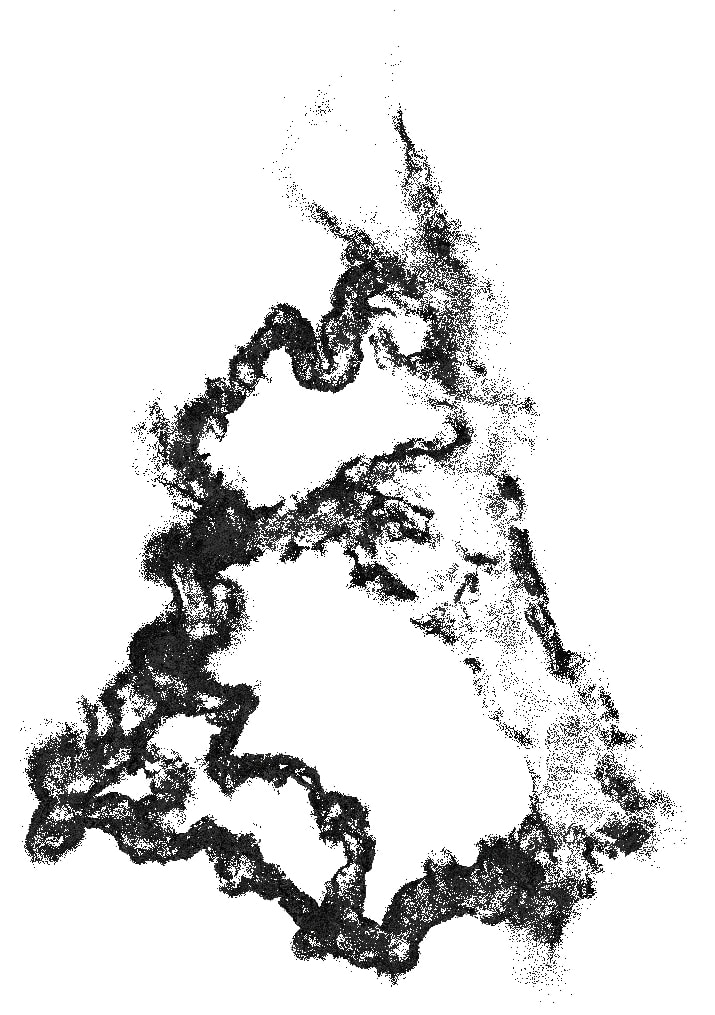}}&
          \hspace{-0.05in}\fbox{\includegraphics[height=0.14\textheight, trim={0in, 0in, 0in, 0in},clip]{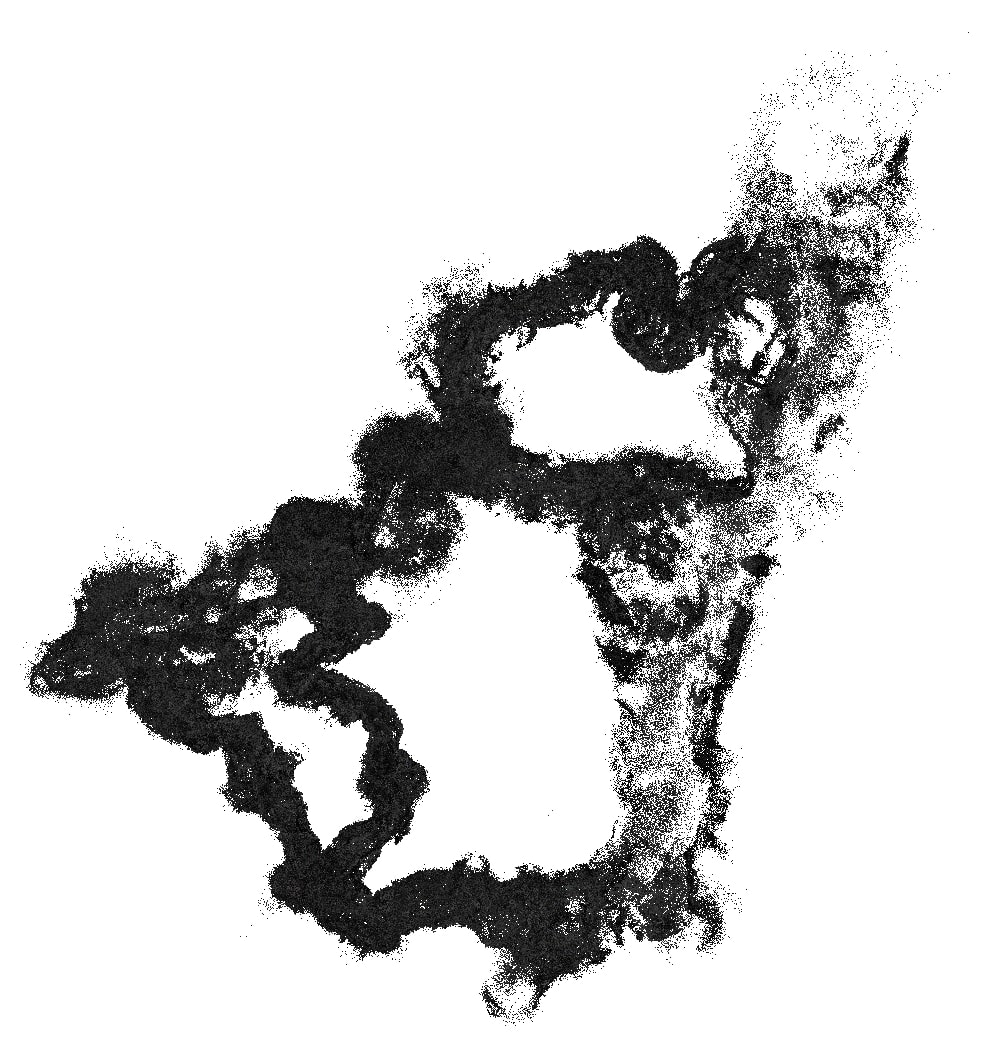}}
   \end{tabular}
    \caption{The left, center, and right GoPro cameras, sparse point cloud produced by SVIn2. On the right, the combined point cloud from all three cameras.}
    \label{fig:sparse}
\end{figure*}
\subsection{Sparse reconstruction, Camera Trajectory Generation, Cave Mapping}
The SVIn2~\cite{RahmanIJRR2022,JoshiICRA2022} package was utilized to extract the camera pose trajectories and sparse point clouds for each of the three cameras (left, center, and right). The un-aligned trajectories can be seen in \fig{fig:TrajMaps} on the left. By utilizing the calibration target placed at the beginning of the catacombs section, the three trajectories are aligned in a common frame of reference and adjusted to the correct depth (z-dimension); see \fig{fig:TrajMaps}. The resulting sparse point cloud representations can be seen in \fig{fig:sparse} for the left, center, and right GoPro respectively, while the combined dataset, aligned in the same coordinate frame can be seen in the right image of \fig{fig:sparse}. Please note that the center camera covers more of the ceiling, while the left and right capture more of the walls. The segment of the main passage near the main line (see \fig{fig:TrajMaps} center) is characterized by a high ceiling and wide passage, which was reflected in the lack of detected points in all three cameras.

\begin{figure*}[h]
    \centering
    \begin{tabular}{lcc}
         \fbox{\includegraphics[angle=90, height=0.18\textheight, trim={0in, 0in, 0in, 0in},clip]{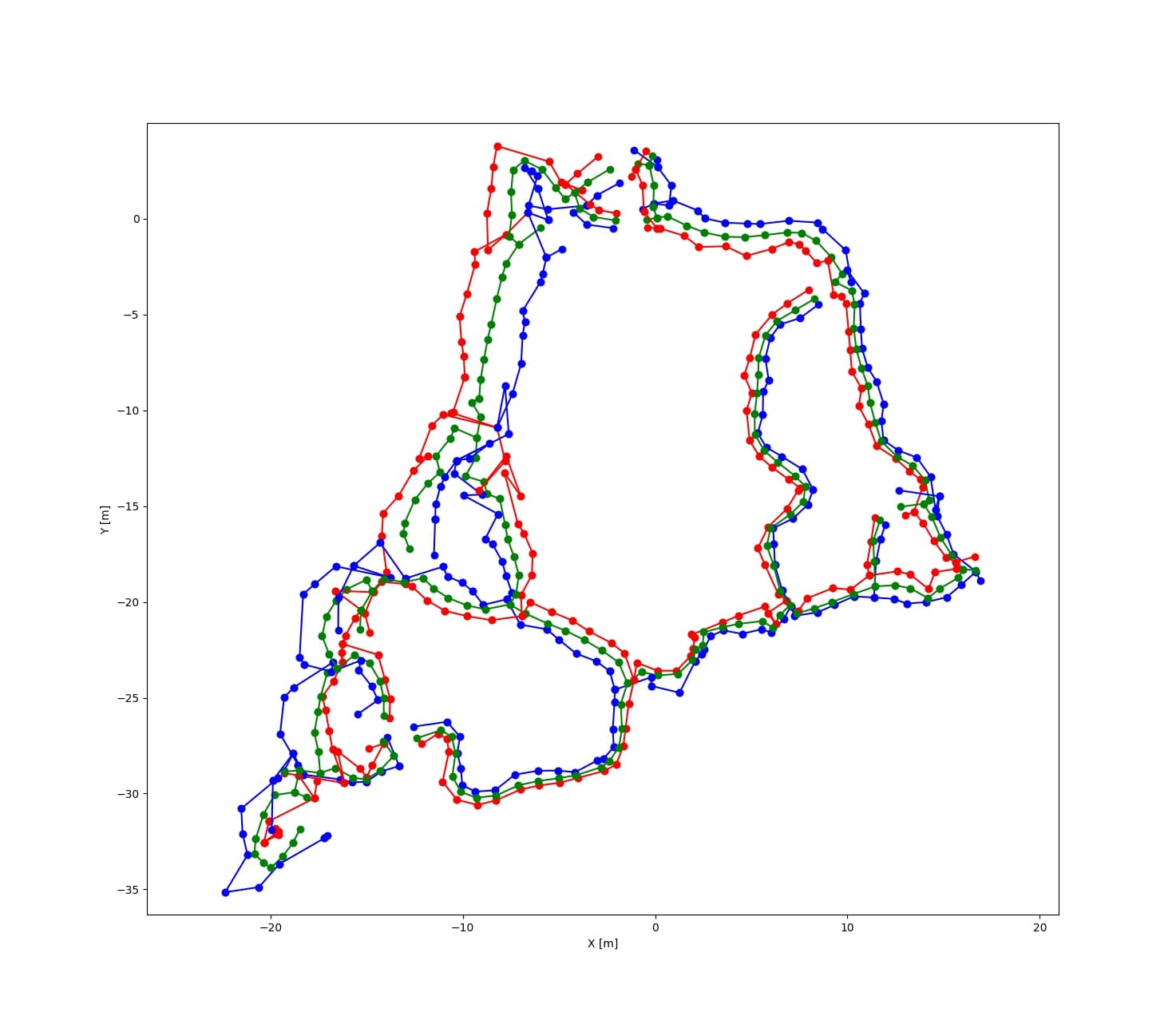}}&
         \fbox{\includegraphics[height=0.18\textheight, trim={0in, 0in, 0in, 0in},clip]{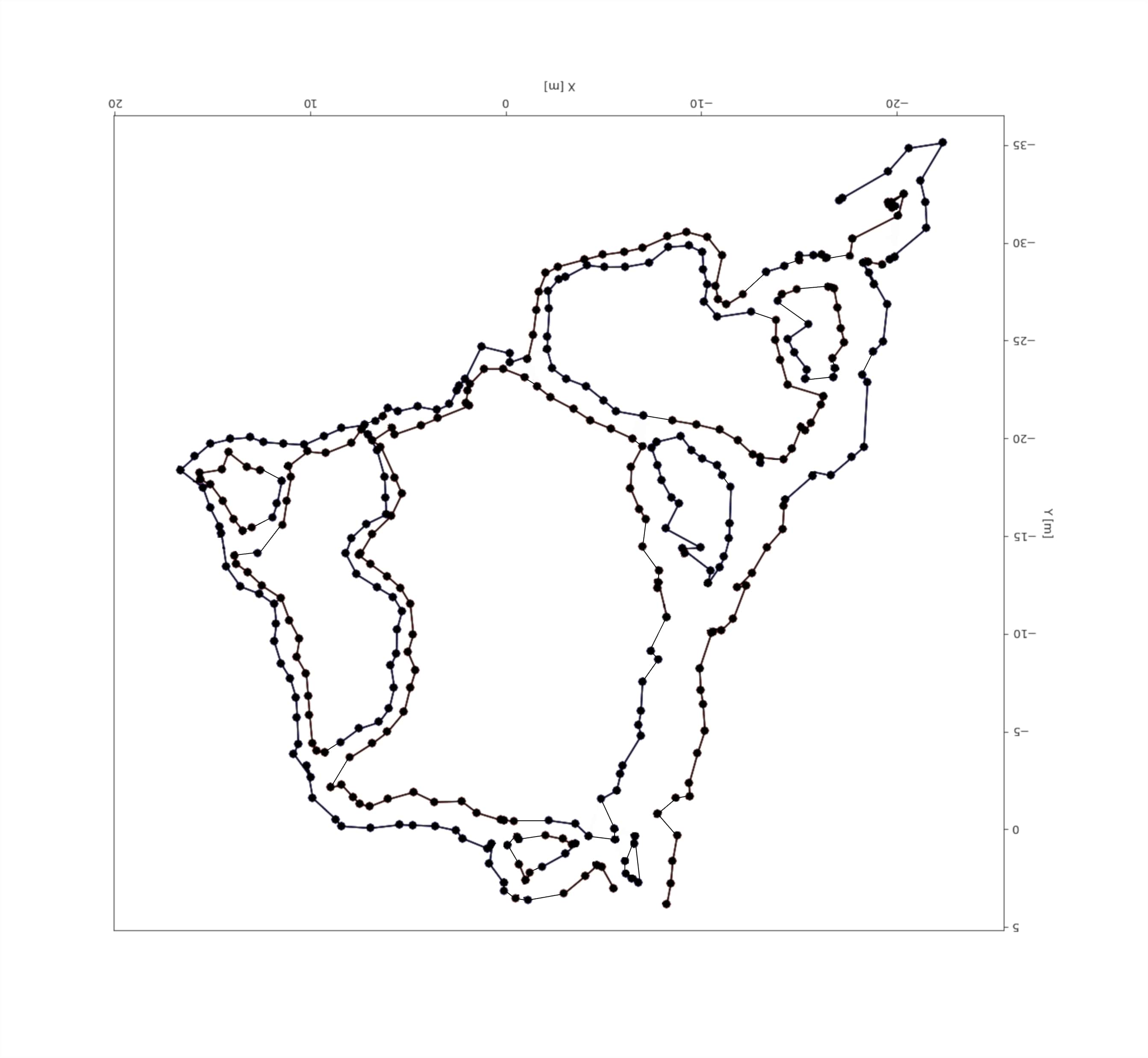}}&
         \fbox{\includegraphics[height=0.18\textheight, trim={0in, 0in, 0in, 0in},clip]{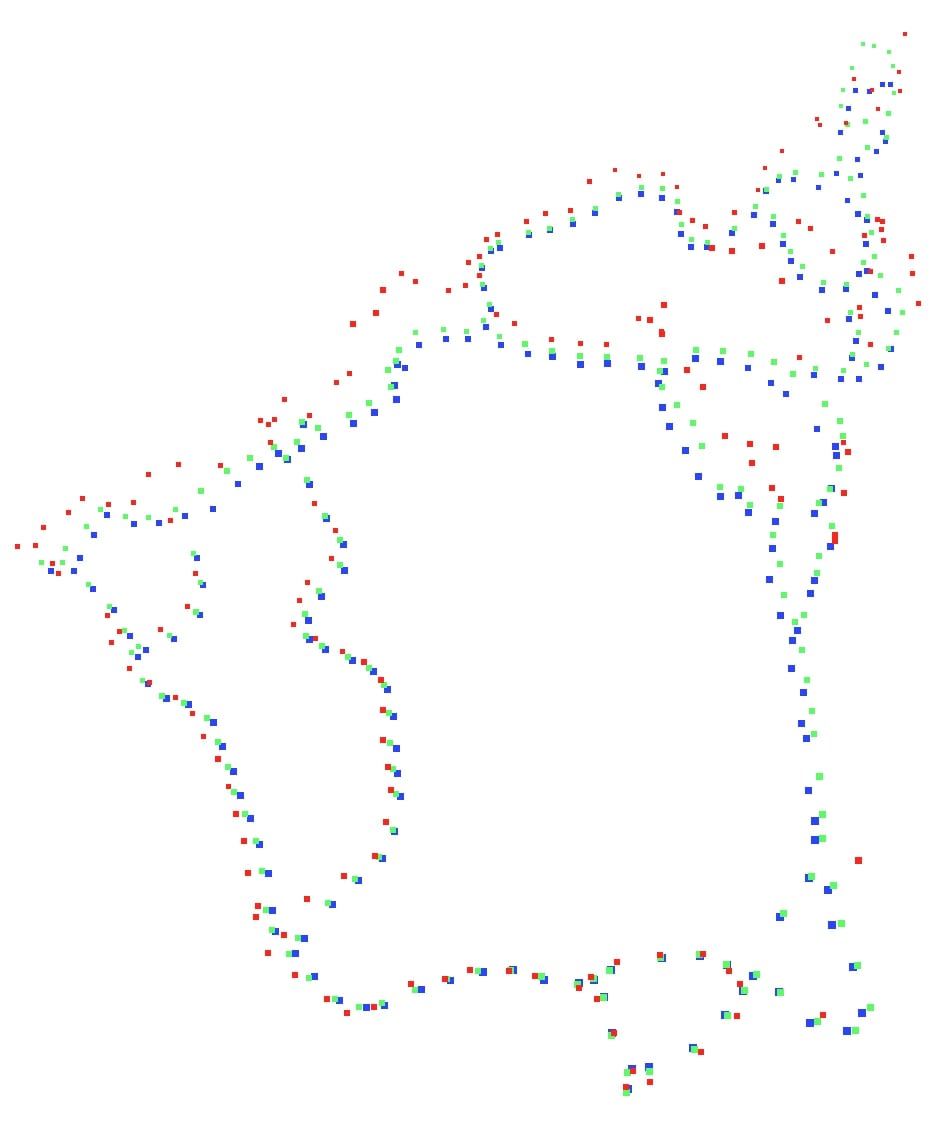}}
   \end{tabular}
    \caption{Left: the average trajectory (in green) connected using the minimum spanning tree (MST) together with the two walls, left (blue) and right (red) extracted from the combined sparse point cloud. Center: the left and right walls (adjusted manually) showing the width of the cave.}
    \label{fig:LRUD}
\end{figure*}

The three trajectories are used as a guideline for the center of the passage. A pose of the center trajectory is selected and every pose in all three trajectories is flagged. The flagged poses are used to calculate an average pose, and then the next pose of the center camera that has not been flagged is selected. The average trajectory produced can in seen in \fig{fig:TrajMaps}. The Minimum Spanning Tree (MST) algorithm is used to connect these average poses and from each one the closest point on the left and right of the average pose (at the same depth -- z-coordinate) is selected. \fig{fig:LRUD} presents the average trajectory in green, the left wall in red and the right wall in blue. As can be seen, this naive method is sensitive in the order of traversal for designating left and right boundaries, and the MST does not capture the full connectivity of the passages. We cleaned up manually the data -- see the center image of \fig{fig:LRUD} -- and the variability in the passage width is captured. Inside the catacombs, the passages are narrower, while at the main line the passage is wide. In addition, the point above and below the average pose (green) are selected, marking the ceiling (red) and the floor (blue)  of the passage; see \fig{fig:LRUD}, where the floor is mapped at an approximately constant distance from the average camera pose while the ceiling displayed greater variability. Over the years, materials brought by the water accumulate on the floor, smoothing it and filling any holes, while the ceiling erodes over time, generating a lot more 3D structure. 

\subsection{Dense reconstruction}
\begin{figure*}[h]
    \centering
    \begin{tabular}{lcc}
         \hspace{-0.05in}\fbox{\includegraphics[height=0.15\textheight, trim={1.0in, 1.0in, 0.9in, 0.9in},clip]{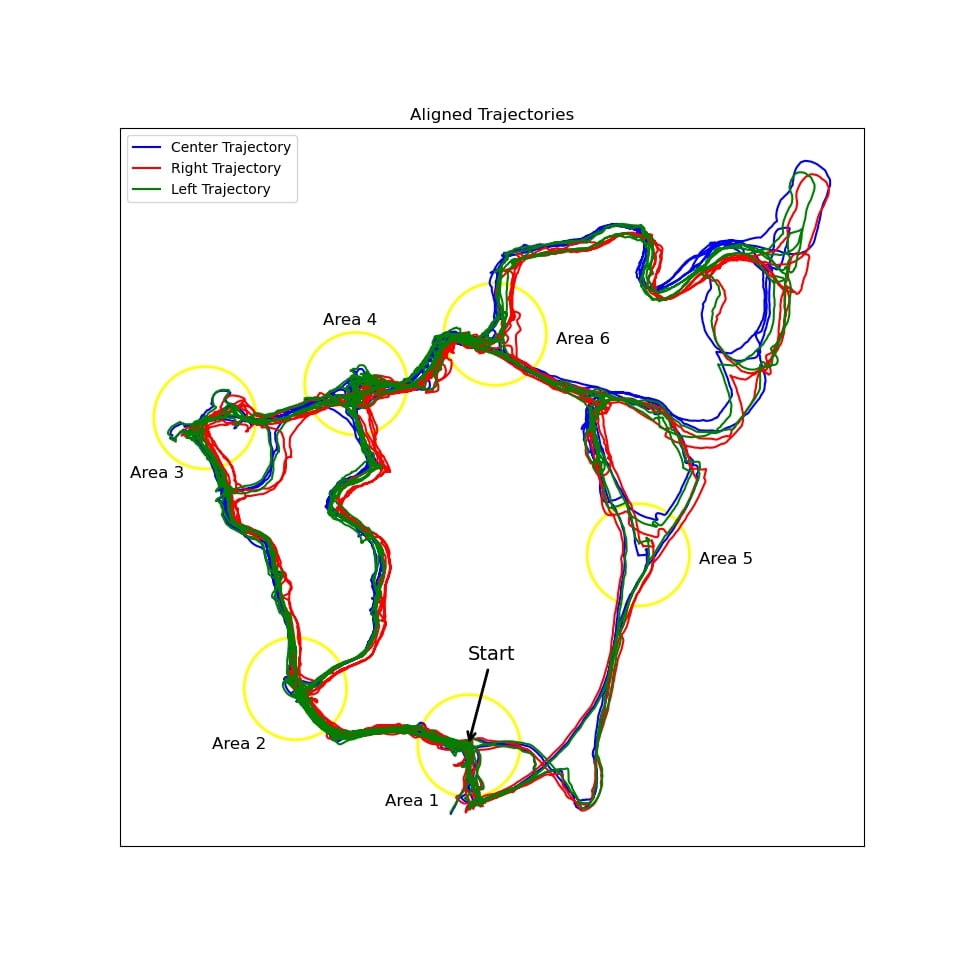}}&
         \hspace{-0.05in}\fbox{\includegraphics[height=0.15\textheight, trim={1.1in, 1.1in, 0.9in, 0.9in},clip]{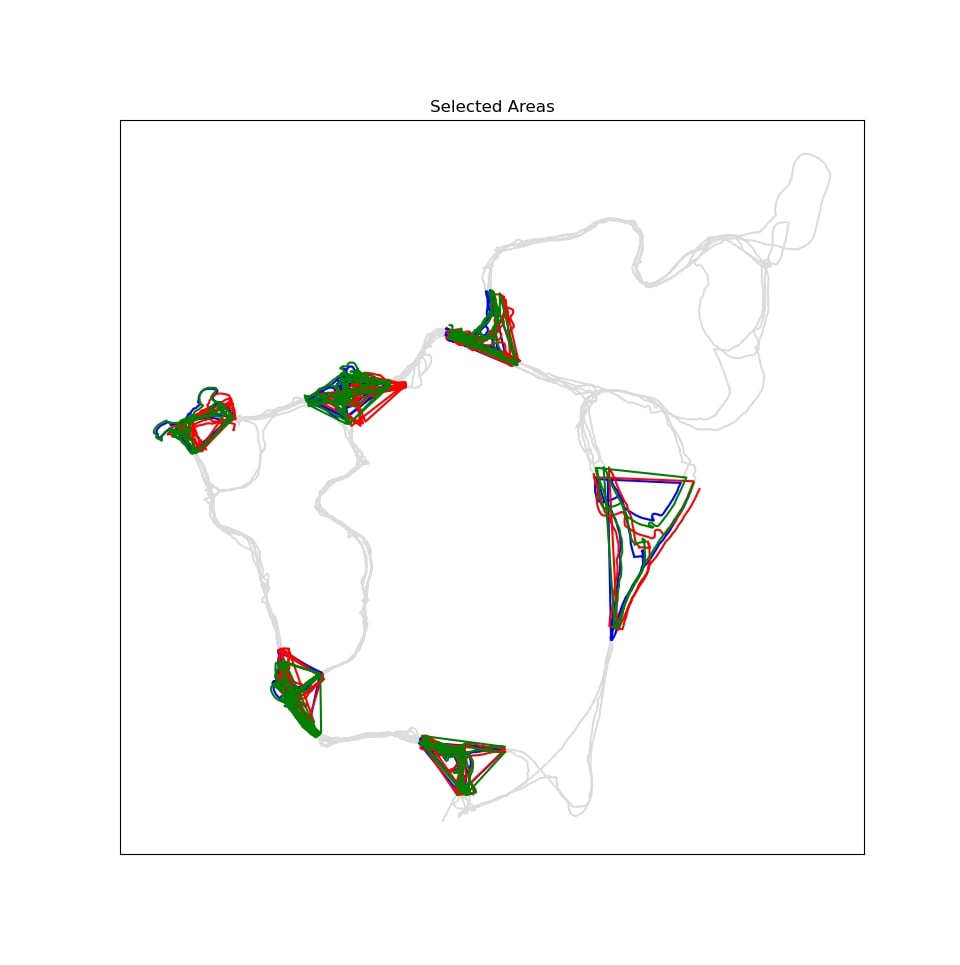}}&
         \hspace{-0.05in}\fbox{\includegraphics[height=0.15\textheight, trim={1.0in, 1.0in, 0.9in, 0.9in},clip]{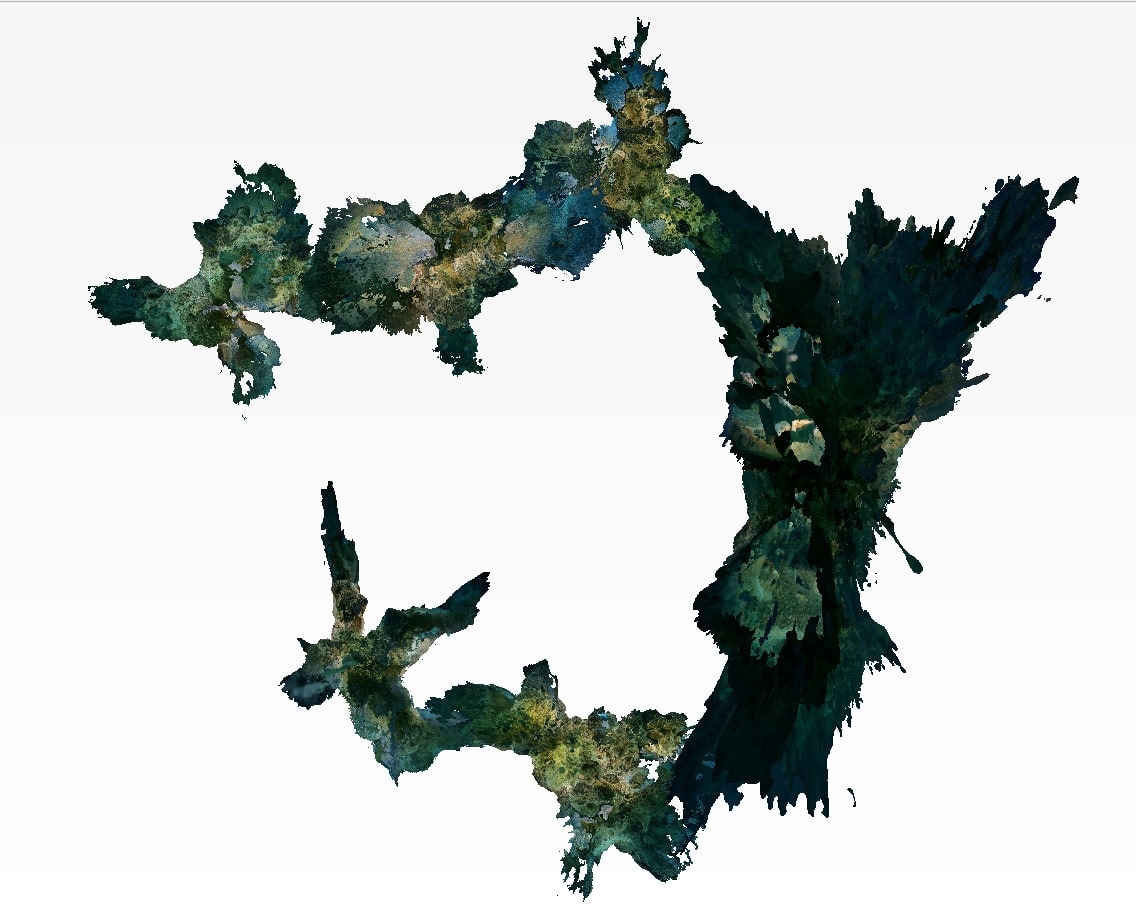}}
    \end{tabular}
    \caption{The aligned trajectories were used as a guideline to select interesting areas for dense reconstruction. On the left, the six areas selected. In the middle, the parts of the trajectories used for the dense reconstruction. On the right, the Poisson reconstruction of the six areas.}
    \label{fig:DenseAreas}
\end{figure*}

\begin{figure*}[h]
    \centering
    \begin{tabular}{lcc}
         \hspace{-0.05in}\fbox{\includegraphics[height=0.12\textheight, trim={0in, 0in, 0in, 0in},clip]{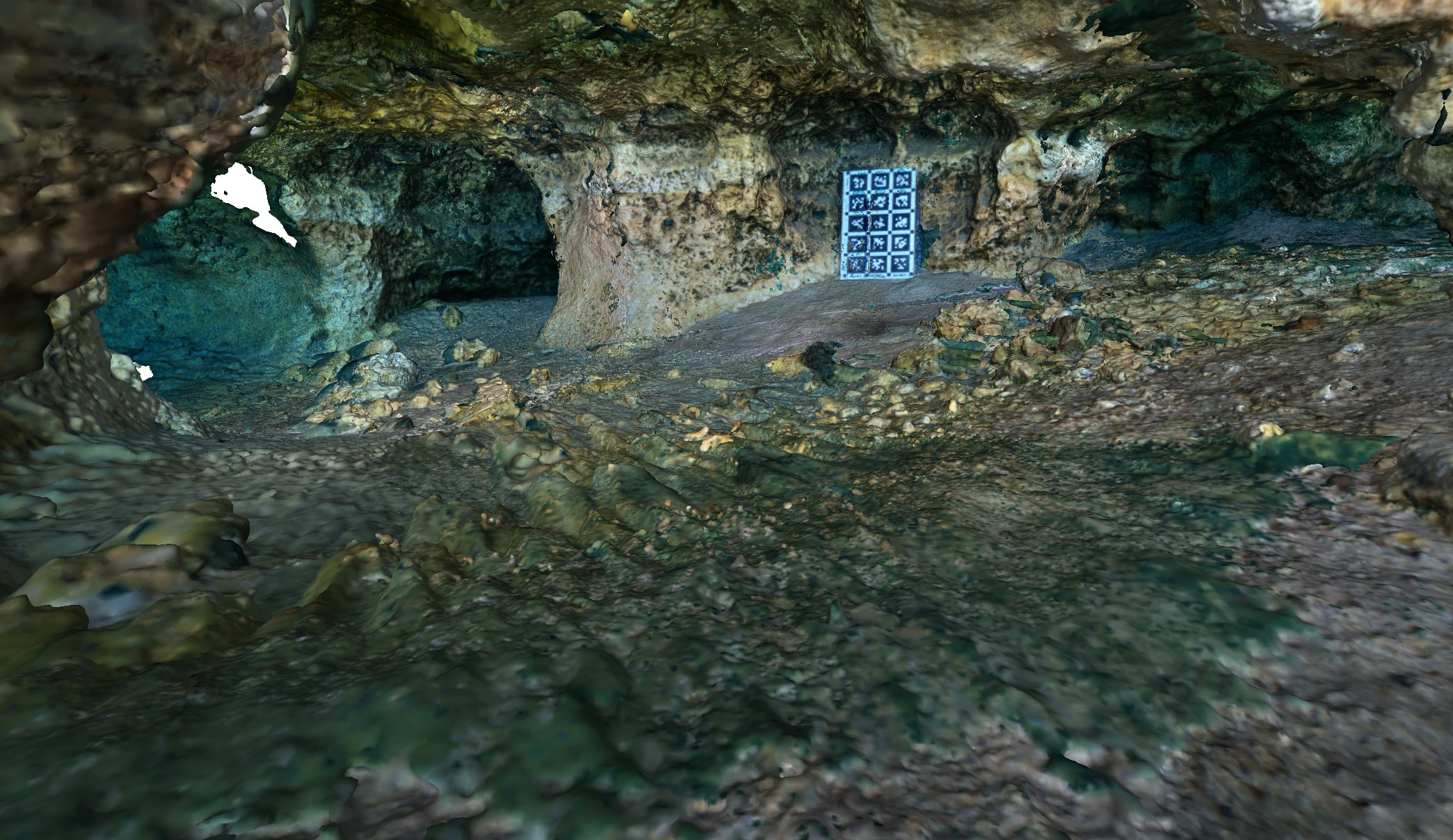}}&
         \hspace{-0.05in}\fbox{\includegraphics[height=0.12\textheight, trim={0in, 0in, 0in, 0in},clip]{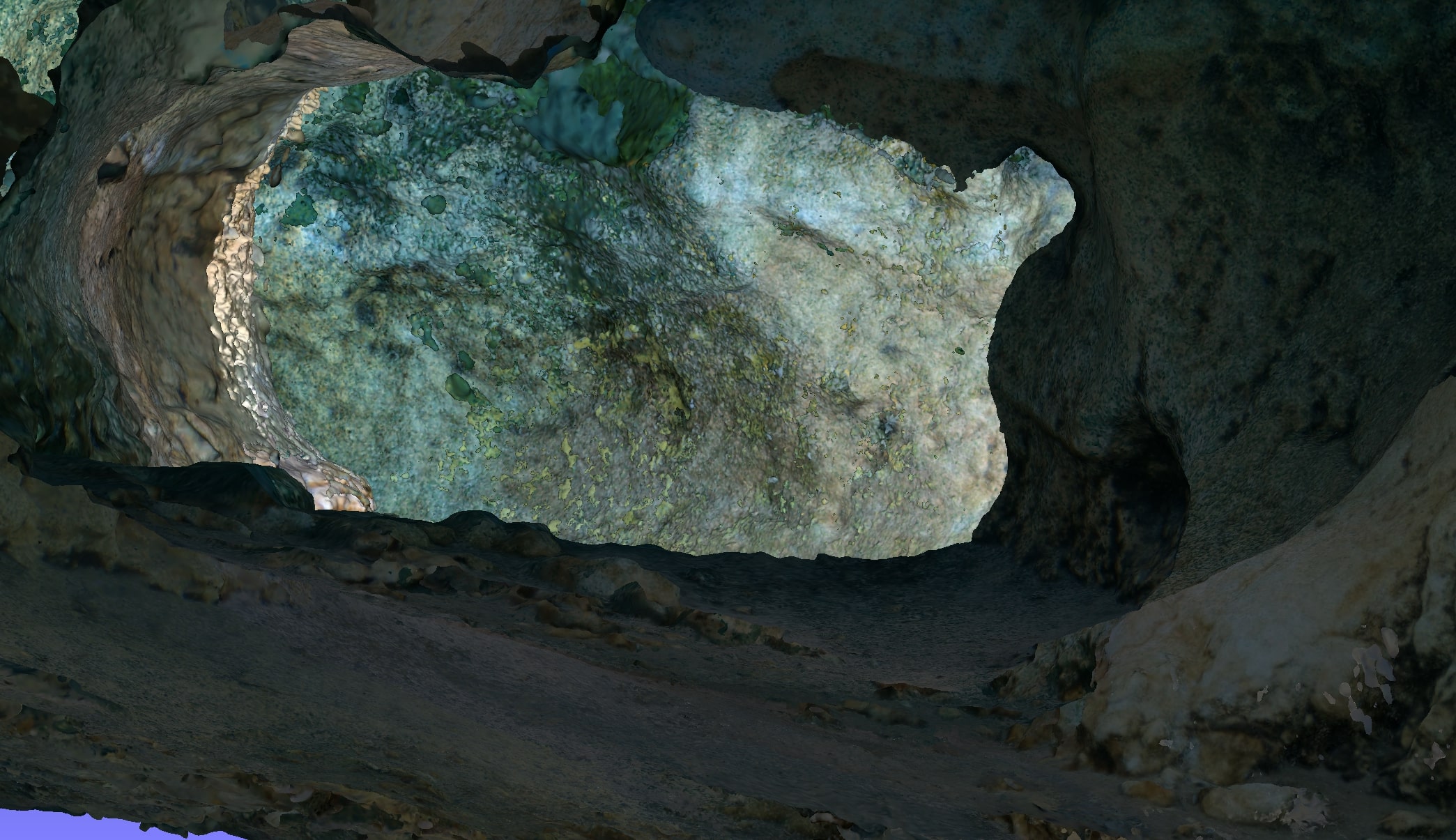}}&
         \hspace{-0.05in}\fbox{\includegraphics[height=0.12\textheight, trim={0in, 0in, 0in, 0in},clip]{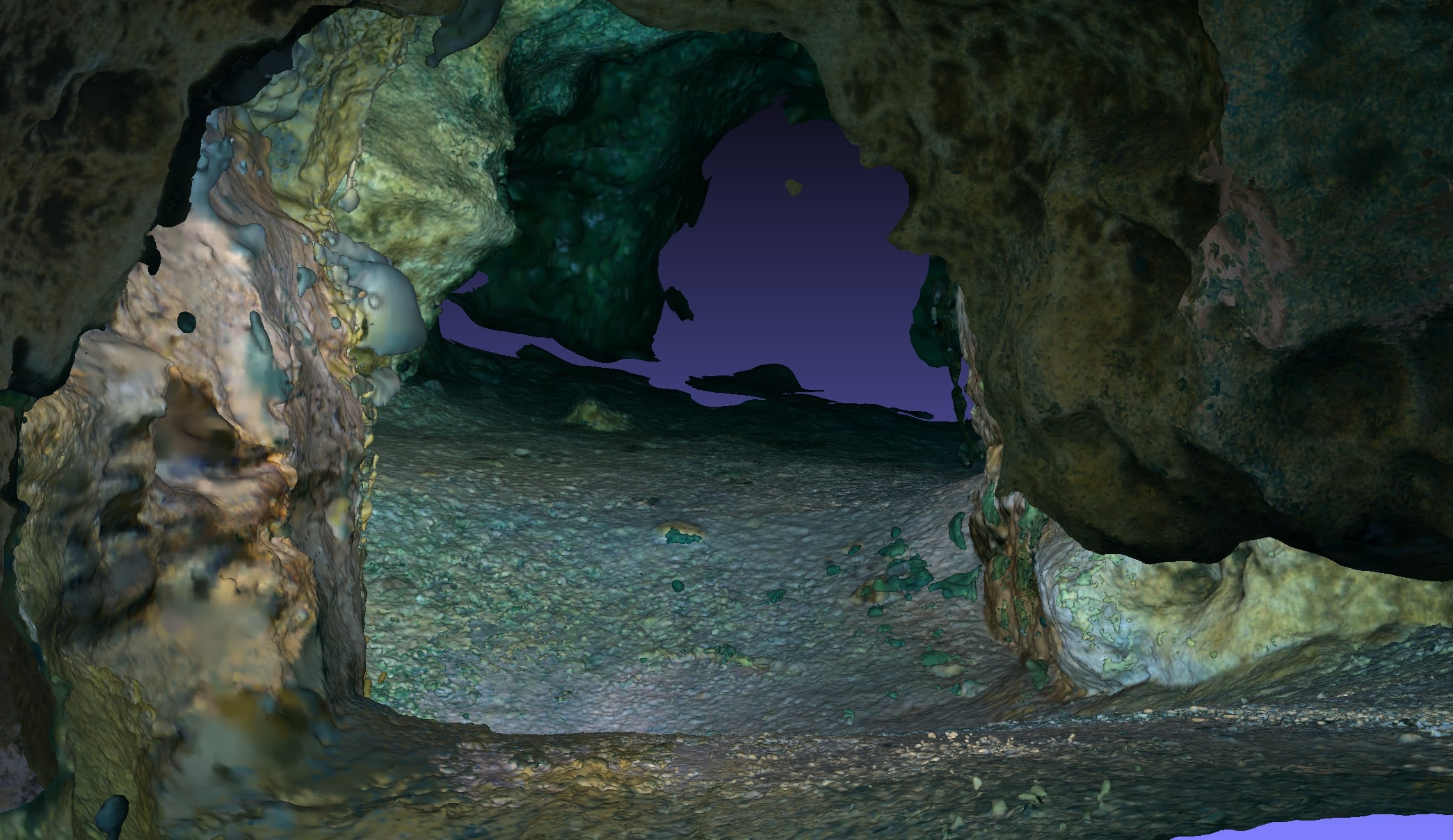}}\\
         \hspace{-0.05in}\fbox{\includegraphics[height=0.12\textheight, trim={0in, 0in, 0in, 0in},clip]{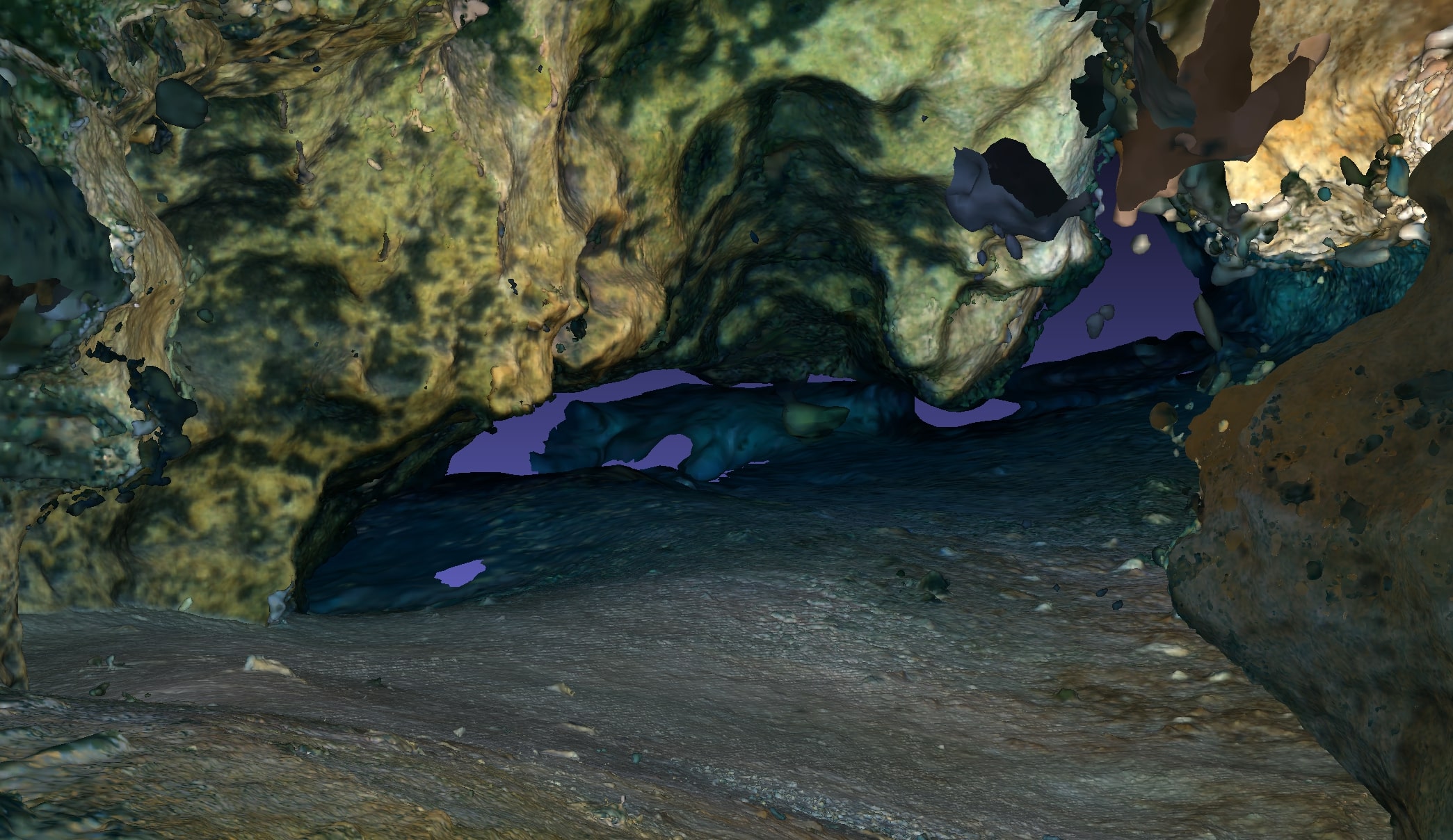}}&
         \hspace{-0.05in}\fbox{\includegraphics[height=0.12\textheight, trim={0in, 0in, 0in, 0in},clip]{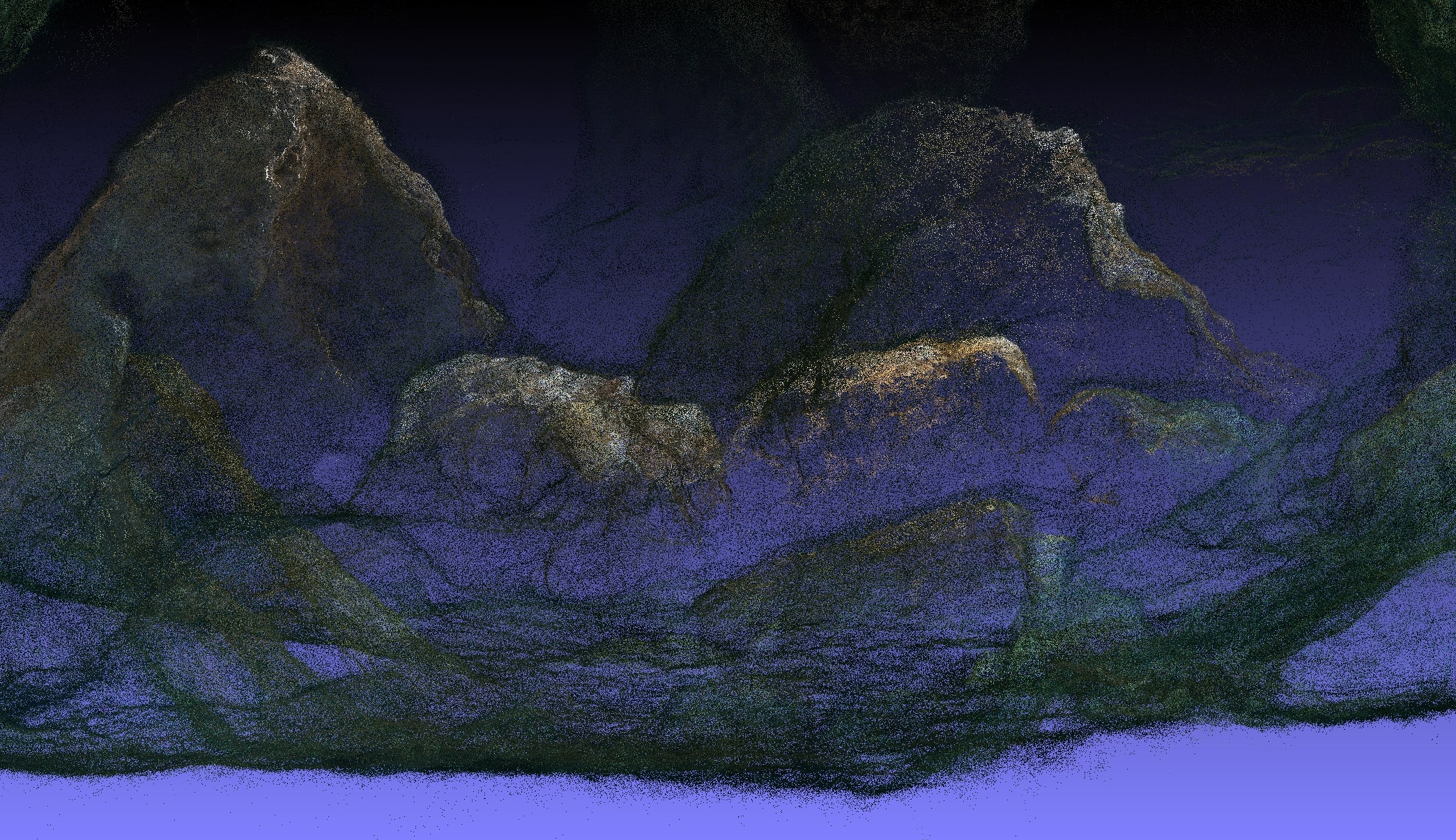}}&
         \hspace{-0.05in}\fbox{\includegraphics[height=0.12\textheight, trim={0in, 0in, 0in, 0in},clip]{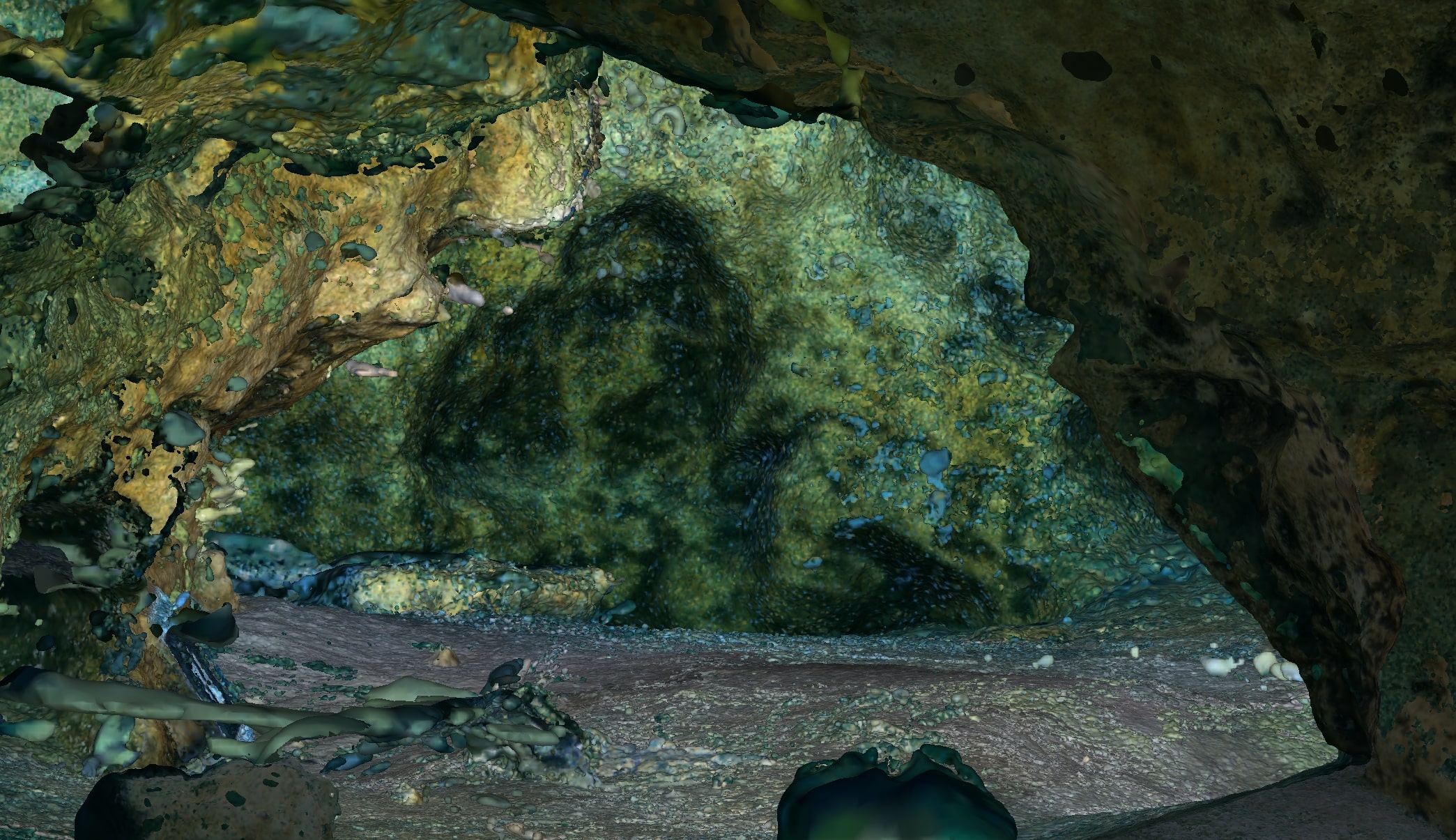}}\\
   \end{tabular}
    \caption{Dense (Poisson) reconstructions from the six areas depicted in \fig{fig:DenseAreas}. Area 1: the start of the recordings, the calibration target is clearly visible. Area 2: inside the catacombs, see the boundaries of the tunnel. Area 3: as the images for the dense reconstruction are limited, the tunnel is interrupted. Area 4: the floor and walls of the passage are clearly visible of special interest are the formations from the ceiling. Area 5: this was an open area with the walls and ceiling of the tunnel too far to be illuminated. Dense reconstruction point cloud of the floor and some rock formations. Area 6: another narrow passage, with a flat floor. }
    \label{fig:denseReco}
\end{figure*}

From the three cameras, more than forty thousand keyframes were generated. In this paper, we selected six representative areas; see \fig{fig:DenseAreas}, for dense reconstruction. For each area, we picked one central pose, taken as the pose for a particular image from the center camera.  We then found all keyframes, from all three cameras, with poses inside a radius of 2.5 meters, as measured from the central pose, except area 5 where a radius of 5 meters was used. Reconstruction was performed using the global optimization package COLMAP~\cite{colmap_sfm}, where the camera poses from SVIn2 were used as constraints, as described in the previous section. From the resulting dense Poisson mesh, representative views are displayed in \fig{fig:denseReco}. More specifically, the calibration target is clearly visible at the right side of the passage, while the entry to the catacombs is visible to the left. Areas 2-4 and 6 are inside the catacombs and are characterized by narrow passages. It is worth noting that the floor is relatively flat, while the ceiling presents a more detailed structure; see Area 4 and 6. Area 5 was selected outside the catacombs, in the main passage of the cave. In this relatively open area, distant points are visible in our keyframes. A limited number of pixels cover an extended area, resulting in a very noisy reconstruction. The dense point cloud captures the rocks that exist in the middle of the main passage, as can be seen in \fig{fig:denseReco}. 

\section{Conclusions and future work}
We presented a framework for mapping underwater caves, utilizing inexpensive action cameras and a dive computer, available to scuba divers. The proposed pipeline based on VI SLAM and SfM allows for sparse and dense reconstruction of parts of the caves on a cave dataset (which will be publicly released) with several branches and loops extending for more than 1km. 

The results showed the ability of the proposed framework to reliably estimate the camera trajectory and achieve dense reconstructions in sections of the cave.

We plan several research directions in the immediate short term based on some of the lessons learned from this paper. 
While the trajectory looks accurate, achieving a dense map of the whole cave system was unattainable, because there were many outliers preventing a good reconstruction. In addition, the occupied memory raised significantly, making it challenging to manage even on a powerful workstation with 256GB. We plan to investigate strategies to iteratively perform dense reconstruction in a window.
We also plan to add to the dataset a 3D sonar released by WaterLinked that has a narrow beam and provides a 3D point cloud.

In the longer term, we plan to integrate the pipeline for real-time 3D reconstruction for eventual robot navigation, towards autonomous mapping of underwater caves.
\bibliographystyle{IEEEtran}
\bibliography{IEEEabrv,refs,pubs,refs_pm}

\end{document}